\documentclass[acmtog, screen, authorversion]{acmart} % arxiv
\acmSubmissionID{278}

\usepackage{booktabs}
\usepackage{times}
\usepackage{epsfig}
\usepackage{graphicx}
\usepackage{amsmath}
\usepackage{amssymb}
\usepackage{overpic}
\usepackage{enumitem}
\usepackage{layouts}
\usepackage{hyphenat,balance,microtype}
\usepackage[fleqn,tbtags]{mathtools}
\usepackage[capitalise]{cleveref} 
\usepackage[detect-weight]{siunitx}

\graphicspath{{./images/}}

\newcommand{\tbf}[1]{\textbf{#1}}
\newcommand{\tablestyle}[2]{\setlength{\tabcolsep}{#1}
                            \renewcommand{\arraystretch}{#2}
                            \centering
                            \footnotesize} % Usage: \tablestyle{4pt}{1.1}

\def \eg {{\emph{e.g}.\thinspace}, }
\def \ie {{\emph{i.e}.\thinspace}, }

\acmYear{2022}
\acmJournal{TOG}
\acmVolume{41}
\acmNumber{4}
\acmArticle{103}
\acmMonth{7} 

\setcopyright{acmcopyright}

\acmDOI{10.1145/3528223.3530087}

\received{January 2022}
\received[accepted]{March 2022}
\received[final version]{May 2022}

\begin{document}

%%%%%%%%% TITLE
\title{Dual Octree Graph Networks for Learning Adaptive Volumetric Shape Representations}

%%%%%%%%% AUTHORS
\author{Peng-Shuai Wang}
\affiliation{
 \institution{Microsoft Research Asia}
 \country China
 }
\email{wangps@hotmail.com}

\author{Yang Liu}
\affiliation{
 \institution{Microsoft Research Asia}
 \country China
 }
\email{yangliu@microsoft.com}

\author{Xin Tong}
\affiliation{
 \institution{Microsoft Research Asia}
 \country China
 }
\email{xtong@microsoft.com}

%%%%%%%%% ABSTRACT
\begin{abstract}
We present an adaptive deep representation of volumetric fields of 3D shapes and an efficient approach to learn this deep representation for high-quality 3D shape reconstruction and auto-encoding. Our method encodes the volumetric field of a 3D shape with an adaptive feature volume organized by an octree and applies a compact multilayer perceptron network for mapping the features to the field value at each 3D position. An encoder-decoder network is designed to learn the adaptive feature volume based on the graph convolutions over the dual graph of octree nodes. The core of our network is a new graph convolution operator defined over a regular grid of features fused from irregular neighboring octree nodes at different levels, which not only reduces the computational and memory cost of the convolutions over irregular neighboring octree nodes, but also improves the performance of feature learning. Our method effectively encodes shape details, enables fast 3D shape reconstruction, and exhibits good generality for modeling 3D shapes out of training categories. We evaluate our method on a set of reconstruction tasks of 3D shapes and scenes and validate its superiority over other existing approaches.
Our code, data, and trained models are available at 
\url{https://wang-ps.github.io/dualocnn}.
%\url{https://github.com/microsoft/DualOctreeGNN}.

\end{abstract}

%%%%%%%%% TEASER
\begin{teaserfigure}
  \centering
  \includegraphics[width=\linewidth]{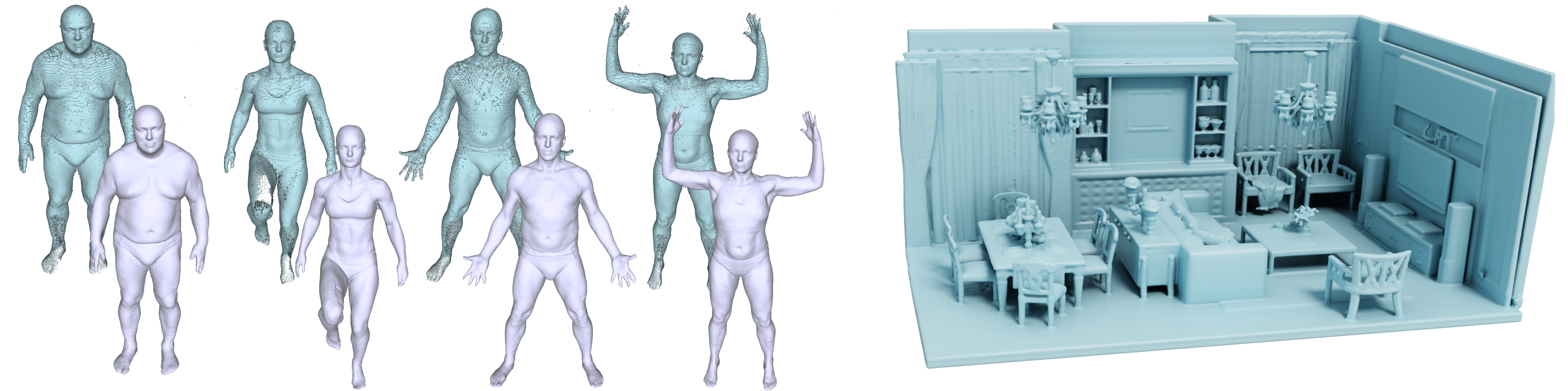}
  \vspace{-6mm}
  \caption{3D reconstruction results generated by our dual octree graph network trained on raw scans of human bodies from the D-Faust dataset~\cite{Bogo2017} via unsupervised learning. Our network directly predicts an adaptive feature volume with a resolution up to $256^3$ from an input point cloud and uses a learned MLP to map the feature volume to the 3D occupancy field of the resulting 3D shapes.
  Left: four unseen point clouds from the testing dataset (top row) and the resulting 3D shapes reconstructed by our network (bottom row).
  Right: the reconstruction result of a complex scene inferred by our network trained on the D-Faust dataset via a single forward pass, which only takes \SI{478}{\ms} on a V100 GPU and clearly demonstrates the efficiency and generality of our method.}
  \label{fig:teaser}
\end{teaserfigure}

%%%%%%%%% CCSXML

% \ccsdesc[300]{Computing methodologies~Point-based models}

% %%%%%%%%% KEYWORDS

%%%%%%%%% BODY TEXT
\maketitle
\section{Introduction} \label{sec:intro}

Volumetric fields of 3D shapes, such as signed distance field (SDF) and occupancy field, have been widely used in 3D modeling and reconstruction ~\cite{Curless1996, Kazhdan2006,Bridson2015} and demonstrated their advantages in recent 3D learning techniques~\cite{Wu2015,Maturana2015,Qi2016,Wu2016,Choy2016,Wang2018b}.
However, developing an efficient deep representation of volumetric fields and the associated learning scheme for 3D shape generation and reconstruction is still an open problem.

Volumetric CNNs directly sample volumetric fields with regular 3D grids.
Although these approaches can easily adapt deep neural networks developed for 2D images to 3D learning, their memory and computational costs grow cubically as the volumetric resolution increases, making them difficult to model 3D shape details.
A set of methods~\cite{Wang2017,Graham2018,Shao2018,Choy2019} represent 3D shapes with sparse non-empty voxels and design neural networks that operate only on sparse voxels.
Although these sparse-voxel-based methods significantly reduce computational and memory cost, the features in empty voxels are ignored or simply set to zero, and predicting the locations of sparse voxels for shape generation and reconstruction is a difficult task, especially for incomplete inputs.
Neural implicit approaches~\cite{Park2019,Mescheder2019,Chen2019,Sitzmann2020,Tancik2020} encode volumetric fields with compact multilayer perceptrons (MLPs) by mapping an input 3D point to a field value and learning the MLP as an autodecoder~\cite{Park2019} when training on a large dataset.
Without efficient encoding networks, the MLP needs to be optimized for each new input at the inference time, which is time-consuming. Moreover, visualizing or extracting 3D shapes from the resulting volumetric field is also expensive due to millions of MLP evaluations for a large number of sampling positions.
Most recently, the work of~\cite{Peng2020} combines regularly-sampled volumetric features with local MLPs to model 3D shapes, which reduces the computational cost of MLP evaluation. However, its ability to model shape details is still limited by the resolution of the feature volume.

% our method.
In this paper, we present an adaptive deep representation of 3D volumetric fields and an efficient learning scheme of this deep representation for 3D shape reconstruction and autoencoding. Our method models the full volumetric field of a 3D shape as an adaptive octree of feature volume and learns an MLP to evaluate the field value at a queried 3D position. The feature at a coarse-level octree node represents the low-frequency variations of the volumetric field around the node, while the feature at the fine-level octree node represents the high-frequency details of the volumetric field around the surface. To generate a continuous volumetric field from the adaptive feature volume, we adapt the multilevel partition of unity (MPU) \cite{Ohtake2003} to interpolate the contributions of the octree nodes surrounding a 3D position, each of which is computed by the MLP with the target 3D position and the node feature as input.

% overall scheme; network.
To learn the adaptive feature volume of 3D shape and the MLP, we design an efficient encoder-decoder network based on graph CNNs that extracts features from the adaptive octree of the input point cloud and determines the subdivision of the feature volume. 
Unlike previous octree-based CNNs \cite{Wang2017,Wang2018a} that compute regular 3D convolution over local octree nodes at the same level, the key of our network is to perform graph convolution over neighboring octree nodes at different levels. For this purpose, we construct the dual octree graph and perform the message-passing graph convolution~\cite{Simonovsky2017,Gilmer2017} over the one-ring neighborhood of each octree node. To reduce the memory and computational cost of irregular graph convolutions at each node, we exploit the semi-regularity of the dual octree graph and develop a regular graph convolution over the features fused from neighboring nodes at six fixed directions, which can be efficiently executed on the GPU.

% advantages.
Our new adaptive deep representation and associated learning scheme have several advantages. The combination of adaptive feature volume and MLP efficiently encodes shape details and enables fast field value evaluation. With the help of new graph convolution operator, our graph network can efficiently learn adaptive feature volume from input point clouds. After training, the encoder-decoder network can directly infer the resulting volumetric field from the input point cloud and exhibits good generality for reconstructing 3D shapes beyond training shape categories.

We evaluate our method in a series of 3D reconstruction tasks including shape and scene reconstruction from noisy point clouds, unsupervised surface reconstruction, and encoding 3D shape collections.
Our method achieves state-of-the-art performance on four benchmarks and is consistently better than existing methods.
Compared with MLP-based methods, our method can generate higher quality results with much less inference time (about 390 times faster than MLP-based methods for surface reconstruction).
Compared with sparse voxel-based methods, our method encodes features of the full volumetric field and is more robust to reconstruct 3D shapes from noisy or incomplete point clouds.
Our ablation studies further validate the efficiency and efficacy of our graph network design and the new graph convolution operator for learning adaptive feature volumes. \looseness=-1

% summary.
In summary, the contributions of our work are
\begin{itemize}[leftmargin=*,nosep]\setlength\itemsep{1mm}
    \item[-] A new adaptive deep representation of 3D volumetric fields for shape reconstruction and autoencoding.
    \item[-] An effective dual octree graph neural network with an encoder-decoder structure that learns multiscale features from the input point cloud and encodes shape details in high quality.
    \item[-] A novel graph convolution scheme that significantly improves both the efficacy of graph convolution and the computational and memory efficiency.
\end{itemize}

\section{Related Work} \label{sec:related}

\paragraph{Volumetric CNNs}
3D volumetric fields can be explicitly represented by discrete values sampled on uniform grids.
Volumetric CNNs directly generalize 2D CNNs to 3D volumetric fields sampled over uniform grids and achieve promising results in 3D shape classification~\cite{Wu2015,Maturana2015,Qi2016}, 
3D generation~\cite{Wu2016,Brock2016}, 3D reconstruction~\cite{Choy2016} and completion~\cite{Dai2017}. 
The main drawback of volumetric CNNs is their high computational and memory cost caused by the cubic complexity of volumetric representation. \looseness=-1

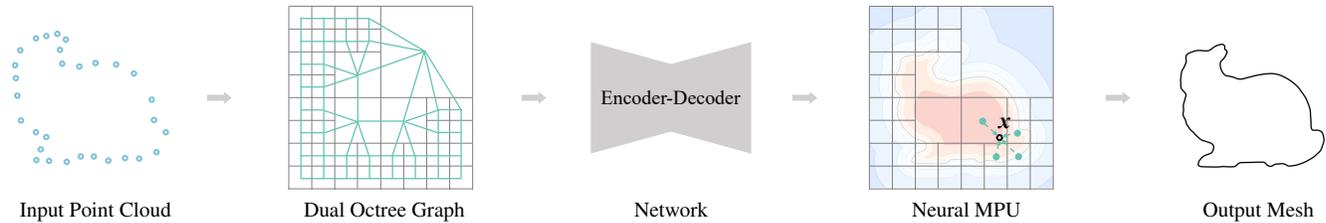
\begin{figure*}[t]
  \centering
   \begin{overpic}[width=0.98\linewidth]{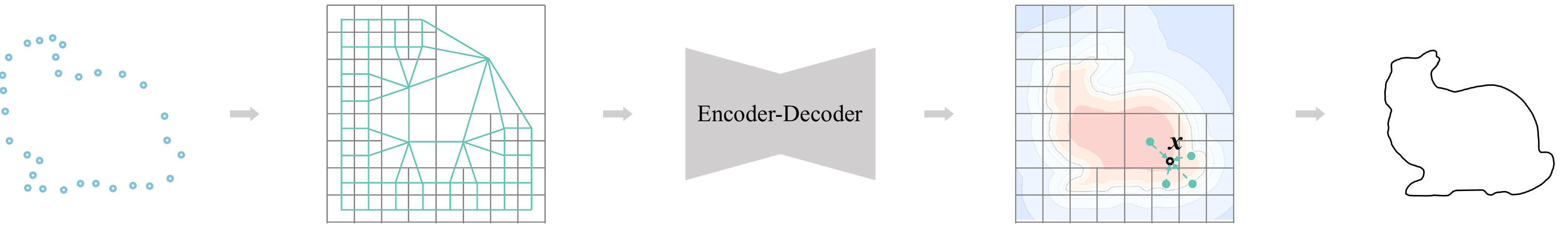}
    \put(0.5, -1){\small Input Point Cloud}
    \put(22, -1){\small Dual Octree Graph}
    \put(47, -1){\small Network}
    \put(68, -1){\small Neural MPU}
    \put(90, -1){\small Output Mesh}
   \end{overpic}
  \caption{An overview of our method illustrated with an 2D example. From left to right: given a point cloud as input, our method first constructs a dual octree graph and then applies a graph-CNN-based encoder-decoder network on the input graph to extract an adaptive feature volume. After that, the learned Neural MPU module maps the adaptive feature volume to a 3D volumetric field of the resulting surface. }
  \label{fig:overview} 
  \vspace{-4mm}
\end{figure*}

\vspace{-4pt}
\paragraph{Sparse-Voxel-based CNNs}
The drawback of volumetric CNNs is overcome by sparse-voxel-based CNNs, which adopt spatially adaptive data structures such as octrees~\cite{Wang2017} and hash tables~\cite{Graham2018,Shao2018,Choy2019} to index non-empty voxels efficiently and constrain the convolution over these sparse voxels. 
A set of sparse-voxel-based CNNs have been proposed for shape reconstruction and generation, where an encoder-decoder network is learned to map input point cloud to octrees with occupancy values~\cite{Tatarchenko2017,Haene2017}, adaptive planar patches~\cite{Wang2018a}, or moving-least-squares points~\cite{Liu2021b}. However, all these approaches require the ground-truth surface for supervision and cannot be used for unsupervised surface reconstruction from point clouds~\cite{Gropp2020}, especially noisy and incomplete ones. 
Different from these sparse-voxel-based methods, our method represents the full volumetric field with an adaptive octree, thus offering a continuous volumetric feature space for 3D learning. The new graph convolution operator introduced in our method is performed over neighboring octree nodes at different levels, which greatly improves reconstruction quality and achieves similar computational and memory efficiency to the 3D convolutions in sparse-voxel CNNs that are always applied over the octree nodes at the same level. Recently, Ummenhofer et al. ~\shortcite{Ummenhofer2021} span 3D convolution to non-empty octree nodes at different levels. Compared to our method, their approach is limited to \emph{balanced} octrees and cannot guarantee the continuity of the resulting volumetric field. \looseness=-1

% \vspace{-10pt}
\paragraph{Coordinate-based MLPs}
A set of methods have been proposed to implicitly model 2D ~\cite{Oechsle2019, Tancik2020,Sitzmann2020} or 3D fields ~\cite{Park2019,Chen2019,Mescheder2019, Mildenhall2020} with coordinate-based MLPs that map each spatial position to its function value. Compared to volumetric-based representations, the coordinate-based MLPs are memory efficient and can preserve the continuity of the resulting 3D volumetric field. Therefore, they have been applied for reconstructing the continuous distance field of 3D shapes from point clouds via unsupervised learning \cite{Atzmon2020,Atzmon2021,Gropp2020} or supervised learning~\cite{Williams2021,Williams2022}. However, large amounts of computations are required to reconstruct high-resolution fields with the coordinate-based MLPs. Moreover, a single MLP is difficult to model 3D shapes with rich details or a large-scale scene.

Local MLPs reduce the computational cost and improve the expressiveness of coordinate-based MLPs by subdividing the whole space into subspaces and model filed function in each subspace with separate MLPs.  
Instant-NGP~\cite{Muller2022} introduces multiresolution Hash tables and optimized GPU code to speed up the MLP training.
A set of methods\cite{Jiang2020,Genova2020,Chabra2020,Chibane2020} uniformly subdivide the 3D space into grids for learning MLPs in each grid. ACORN~\cite{Martel2021}, NGLOD~\cite{Takikawa2021}, and OctField~\cite{Tang2021} partition the 3D space with octrees. Without efficient encoders, most of these methods can only be used for fitting a single 3D shape or image. Similar to our method, 
OctField ~\cite{Tang2021} develops an encoder-decoder network for surface reconstruction. However, it follows sparse-voxel-based methods ~\cite{Liu2021b,Wang2018a} and only encodes 3D fields on non-empty octree nodes, and the 3D convolution in their method is executed on octants or voxels at the same level. The volumetric field inferred by OctField may not be continuous and thus lead to artifacts in the reconstructed 3D surface. On the contrary, our method models the full volumetric field and guarantees the continuity of the resulting volumetric field. Our graph CNN network is performed on a graph of all neighboring octants at different levels, which effectively improves the expressiveness of the deep representation learned by our method.
ConvONet~\cite{Peng2020} combines volumetric CNNs and MLPs and designs an encoder-decoder network for learning deep representation and MLPs. Because its feature volume is built upon low-resolution uniform grids, the ConvONet cannot faithfully model the geometric details as other adaptive schemes, such as ACORN. Our method benefits from both adaptive feature volumes and an efficient encoder-decoder network. As a result, it exhibits strong expressiveness like ACORN and good generality with the encoder learned from large-scale data.

% \vspace{-10pt}
\paragraph{Graph CNNs for 3D Learning}
Graph CNNs have been developed for graphs of irregular data and have broad applications in many fields (\emph{cf}.\thinspace survey~\cite{Wu2020}).
For 3D learning, various graph CNN schemes have been developed for irregular 3D point clouds. These works usually follow the message passing framework~\cite{Gilmer2017,Simonovsky2017} built upon a k-nearest-neighbor graph of the unorganized point cloud. 
PointNet++~\cite{Qi2017} aggregates the features by a local PointNet~\cite{Qi2017a}. 
NNConv~\cite{Simonovsky2017} defines the local weight function as an MLP that maps the relative position of neighboring points to the corresponding weights.
SpiderCNN~\cite{Xu2018} defines the convolution kernel as a product of a step function and a Taylor polynomial.
SplineConv~\cite{Fey2018} and KPConv~\cite{Thomas2019} define the weight function as a continuous spline function.
PAConv~\cite{Xu2021} builds the convolution kernel by dynamically gathering the weights stored in a memory bank based on the local point positions.
Different from these methods, our graph convolution operates on dual octree graphs of an input point cloud. We design a new graph convolution operator that exploits the semi-regularity of dual octree graphs to improve the efficiency and efficacy of graph CNNs for our task.

\paragraph{Dual Octrees}
Dual octrees have been used in isosurface extraction~\cite{Ju2002,Schaefer2004} and surface reconstruction~\cite{Sharf2007}.
The most widely used dual octree graph extraction algorithm is the recursive procedure proposed by~\cite{Ju2002}, and the follow-up works speed up the process with the help of extra hash tables~\cite{Leon2008,Lewiner2010} on the GPU.
In this paper, we propose a simple-yet-efficient GPU-based scheme to simultaneously construct multiple-level dual octree graphs, which are required for our deep learning tasks.  
%  for accelerating the dual octree construction during network training. 
%  To the best of our knowledge, our approach is the first 3D deep learning method based on dual octree graphs.

\section{Dual Octree Graph Networks} \label{sec:method}

\paragraph{Overview} 
As illustrated in \cref{fig:overview}, we design an encoder-decoder neural network based on dual octree graphs for 3D shape reconstruction and autoencoding. Given a point cloud (possibly incomplete and noisy) sampled from a 3D shape, our method first constructs a dual octree graph of the point cloud (\cref{subsec:dgraph}). An encoder-decoder neural network then takes the dual octree graph as input to predict an adaptive feature volume of the 3D shape (\cref{subsec:docnn}) . After that, a multilevel neural partition-of-unity (Neural MPU) module maps the feature volume to a continuous volumetric field (e.g. signed distance field or occupancy field) for surface extraction (\cref{subsec:mpu}). In the following part of this section, we first introduce the key components of our method described above and then describe the detailed network architecture and loss functions used in training in \cref{subsec:details}.

\begin{figure}[t]
  \centering
  \begin{overpic}[width=\linewidth]{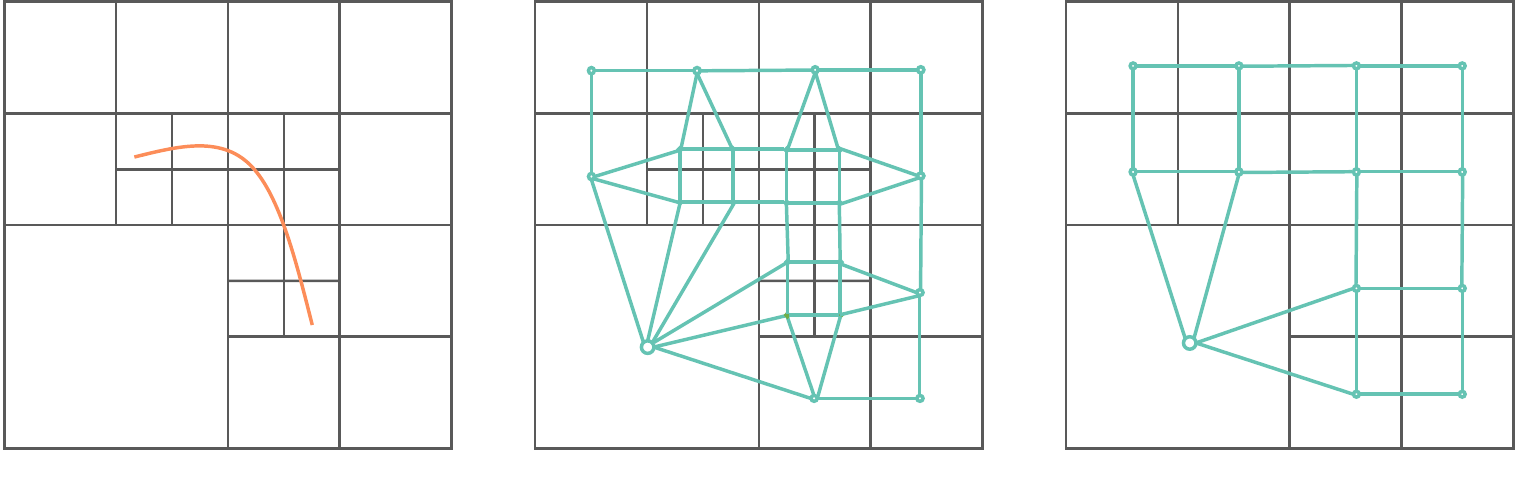}
    \put( 9, -0.5){\small (a) Octree}
    \put(46, -0.5){\small (b) $\mathcal{G}^3$}
    \put(80, -0.5){\small (c) $\mathcal{G}^2$}
  \end{overpic}
  \caption{A 2D illustration of dual octree graph (using quadtree). (a): an octree built from the points sampled from the orange curve. (b)\&(c): the dual octree graphs of the octree under different resolutions, the green lines in the figure are graph edges.
  }
  \label{fig:dgraph} 
  \vspace{-4mm}
\end{figure}

\subsection{Dual Octree Graph} \label{subsec:dgraph}

\paragraph{Dual Octree Graph}
Given a 3D shape, an octree can be built by recursively subdividing the octree nodes whose occupied space intersect with the shape, until the specified maximal tree depth is reached.
The octree allocates finer voxels near the surface and coarser voxels away from the surface. The leaf nodes of an octree form a complete partition of the 3D volume.
The dual octree graph of the octree is formed by connecting all face-adjacent octree leaf nodes with different resolutions.
Formally, we denote an octree with maximum depth $d$ by $\mathcal{T}^d$, and the corresponding \emph{dual octree graph} by $\mathcal{G}^d= \{\mathcal{V}^d, \mathcal{E}^d\}$, where $\mathcal{V}^d = \{v_i\}$ and $\mathcal{E}^d = \{e_{ij}\}$ represent the graph vertex list and the graph edge list, respectively.
A vertex $v_i$ can be identified by $(x_i, y_i, z_i, d_i)$, where the first 3 channels represent the coordinates of the bottom-left corner of a leaf octree node, and the last channel represents the depth of the corresponding octree leaf node.
Two vertices $v_i$ and $v_j$ form an edge $e_{ij} \in \mathcal{E}^d $ if the facets of their corresponding voxel nodes are adjacent.  
We also let $\mathcal{T}^k$ denote the subtree of $\mathcal{T}^d$ constructed by removing the nodes whose depths are greater than $k$, and  $\mathcal{G}^k$ as the dual octree graph of $\mathcal{T}^k$.
A 2D illustration of dual octrees is provided in \cref{fig:dgraph}.

\paragraph{Dual Octree Graph Construction}
Given a 3D point cloud and the maximal octree depth $D$, we need to construct its octree $\mathcal{T}^D$ and dual octree graphs under different resolutions $\mathcal{G}^k, k=1, 2, \ldots, D$. Here, the multi-resolution dual octree graphs are required by the hierarchical encoder and decoder in our network.

For an octree $\mathcal{T}^D$, we employ the algorithm of \cite{Zhou2011} to parallelize the construction process, and we force the octree to be full when the octree depth $d \leq 3$.
We design a simple-yet-efficient algorithm to build the dual octree hierarchy progressively.
Note that when the octree is full, the octree degenerates to uniform voxel grids and the dual graph can be built trivially.
Given a dual graph $\mathcal{G}^d = \{\mathcal{V}^d, \mathcal{E}^d\}$ and a list $\mathcal{L}^d$ indicating which of the nodes in $\mathcal{V}^d$ is subdivided, we can build the dual graph $\mathcal{G}^{d+1} = \{\mathcal{V}^{d+1}, \mathcal{E}^{d+1}\}$ as follows. \looseness=-1

\begin{figure}[t]
  \centering
  \begin{overpic}[width=\linewidth]{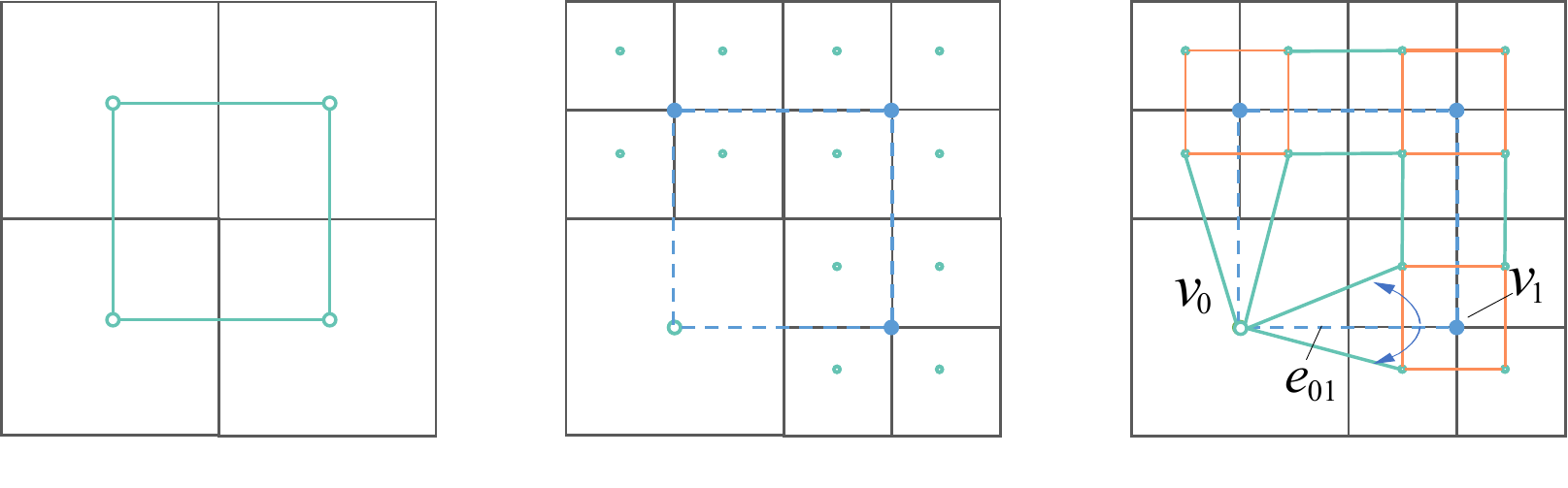}
    \put( 2, -1.5){\small (a) $\mathcal{G}^d = \{\mathcal{V}^d, \mathcal{E}^d\}$}
    \put(42, -1.5){\small (b) Build $\mathcal{V}^{d+1}$}
    \put(76, -1.5){\small (c) Build $\mathcal{E}^{d+1}$}
  \end{overpic}
  \caption{Progressive construction of a dual octree graph. 
  (a): the initial dual octree graph $\mathcal{G}^d$. 
  (b): the blue vertices are invalid and their children are added to the vertex list $\mathcal{V}^d$ for constructing $\mathcal{V}^{d+1}$.
  (c): the dashed blue edges are replaced with 2 new edges, and the orange edges between sibling octree nodes are added via a lookup table.}
  \label{fig:build_graph}
  \vspace{-4mm}
\end{figure}

\begin{itemize} [leftmargin=10pt,itemsep=2pt]
  \item[-] Initialize $\mathcal{V}^{d+1}$ as $\mathcal{V}^d$. According to $\mathcal{L}^d$, we first mark all subdivided octree nodes as \emph{invalid}, then remove them and add their child nodes to $\mathcal{V}^{d+1}$. The step is depicted in \cref{fig:build_graph}-(b).
  \item[-] Initialize $\mathcal{E}^{d+1}$ as $\mathcal{E}^d$. We first mark all edges that connect invalid vertices as invalid (see dashed blue edges in \cref{fig:build_graph}-(b)\&(c)), then replace these edges according to their edge directions. For example, since the edge $e_{01}$ in \cref{fig:build_graph} is a horizontal edge and node $v_1$ is invalid, we replace $v_1$ with its two child nodes on its left side, thus producing two new edges. Finally, we add edges that connect the eight sibling octree nodes belonging to the same parent, which can be simply implemented via a fixed lookup table.
\end{itemize}
The above algorithm can be easily implemented with PyTorch and executes efficiently on CPU or GPU.
The time complexity of building an octree is $\mathcal{O}(N log N)$ on CPU, where $N$ is the number of input points, and the time complexity of building a dual octree graph increases linearly with the number of octree nodes.
For a point cloud with \num{3000} points, the pre-processing time to build an octree and its dual graph is about \SI{2}{\ms} on an Intel I7 CPU.
However, previous methods for building dual octrees either are based on recursive functions~\cite{Ju2002} which is hard to parallelize, or require additional hash tables~\cite{Lewiner2010} to facilitate the building process. \looseness=-1

\paragraph{Dual Octree Graph in Encoder-Decoder Networks}
In the encoder part of the network, the octree is built from the input shape and the node splitting information $\mathcal{L}^d$ is stored in the octree structure.
We directly follow the procedure above to build hierarchical dual graphs in the pre-processing stage.
In the decoder part of the network, the network may predict and grow the octree dynamically for applications where the octree subdivision status is unknown in prior, \eg shape completion and generation.
We follow Adaptive O-CNN~\cite{Wang2018a} and train a predictor which is composed of 2 fully connected layers (FC) to predict whether the current octree node should be subdivided or not.
These predictors can be learned with the ground-truth octree during the training stage. 
In the testing stage,  the node subdivision status $\mathcal{L}^d$ is predicted and the dual graph is updated accordingly with the algorithm above.

\subsection{Dual Octree Graph CNNs} \label{subsec:docnn}

Our network operates on hierarchical dual octree graphs. The main neural network operations include graph convolution (see \cref{subsec:gconv}), graph downsampling and upsampling, and skip connections for a U-Net architecture (see \cref{subsec:sample}).

\subsubsection{Dual octree graph convolution} \label{subsec:gconv}

\paragraph{Graph Convolution}
A graph convolution operation takes a graph  $\mathcal{G} = \{\mathcal{V}, \mathcal{E}\}$ and the node features $F=\{F_i\}$ as input, aggregates and updates $F$ according to the neighborhood information and local node features. Graph convolutions under the message passing or local aggregation framework~\cite{Gilmer2017,Simonovsky2017,Battaglia2018} can be formulated as:
\begin{equation}
  F_i = \mathcal{A} \circ \{ W(\Delta p_{ij}) \times  F_j, \forall j \in \mathcal{N}_i\},
  \label{equ:msg}
\end{equation}
where $\mathcal{A}$ is a differentiable aggregation operator (e.g., \textsc{sum}, \textsc{mean} or \textsc{max}), $W(\cdot)$ is a function that maps the relative position $\Delta p_{ij}$ of two nodes connected by an edge to the convolution weights, $\mathcal{N}_i$ is the neighborhood of $v_i$.

\paragraph{Dual Octree Graph Convolution}
A dual octree graph possesses nice properties for designing efficient graph convolution operators.

First, a dual octree graph is semi-regular: the vector $\Delta p_{ij}$ has only a finite number of values. Thus, for \cref{equ:msg}, no matter how $W(\cdot)$ is defined, we can only get a finite number of $W(\Delta p_{ij})$. Therefore, we set $W(\cdot)$ directly as a weight matrix $W=(W_1,\ldots, W_k)$ where $k$ is the row number, and retrieve the corresponding weights of $\Delta p_{ij}$ by matrix indexing. Compared with previous methods that define $W(\cdot)$ as a continuous function using MLPs~\cite{Wu2019a,Simonovsky2017}, our $W$ is simpler and more efficient. $W$ is also similar to the image convolution weight kernels, and during the training process, each entry of our $W$ can be optimized independently, endowing our graph convolution with strong expressiveness.
 
Secondly, the neighborhood octree nodes of each node can be ordered according to the relative directions from the neighbors to the node.
A dual octree node with the largest depth has exactly 6 neighbors in $+x, -x, +y, -y, +z$ and $-z$ directions, and the majority of dual octree nodes are of this type since the number of leaf nodes of an octree decreases exponentially along with the node depth.
Hence, we set the row number of $W$ to 7 (with the node itself included), \ie mapping $\Delta p_{ij}$ to $\{0, 1, \ldots, 6\}$ according to its relative direction.
For nodes with a smaller depth, although there may be multiple neighboring nodes in each coordinate direction, we use the same weight for these neighboring nodes in the same direction when doing convolutions to achieve simplicity and efficiency. 

Lastly, the node in our graph is a voxel with a non-zero volume, not a simple sample point. The volume size of the node is determined by its octree depth $d_i$. To distinguish nodes at different depth levels, we convert the node depth $d_j$ of $v_j$ to a one-hot vector $D_j$ and concatenate it to $F_j$ as an augmented node feature. We further concatenate $\Delta p_{ij}$ to $F_j$ to increase the expressiveness.

In summary, our dual octree graph convolution is defined as
\begin{equation}
  F_i = \sum_{j \in \mathcal{N}_i } W_{\mathcal{I}(\Delta p_{ij})} \times 
                          [F_j \mathbin\Vert D_j \mathbin\Vert \Delta p_{ij}],
  \label{equ:conv}
\end{equation}
where $\mathcal{I}(\cdot)$ defines an operator that returns the index ( $ \in \{0, 1, \ldots, 6\}$) of a relative direction of $\Delta p_{ij}$, and $\mathbin\Vert$ is the concatenation operator.

\begin{figure}[t]
  \centering
  \begin{overpic}[width=\linewidth]{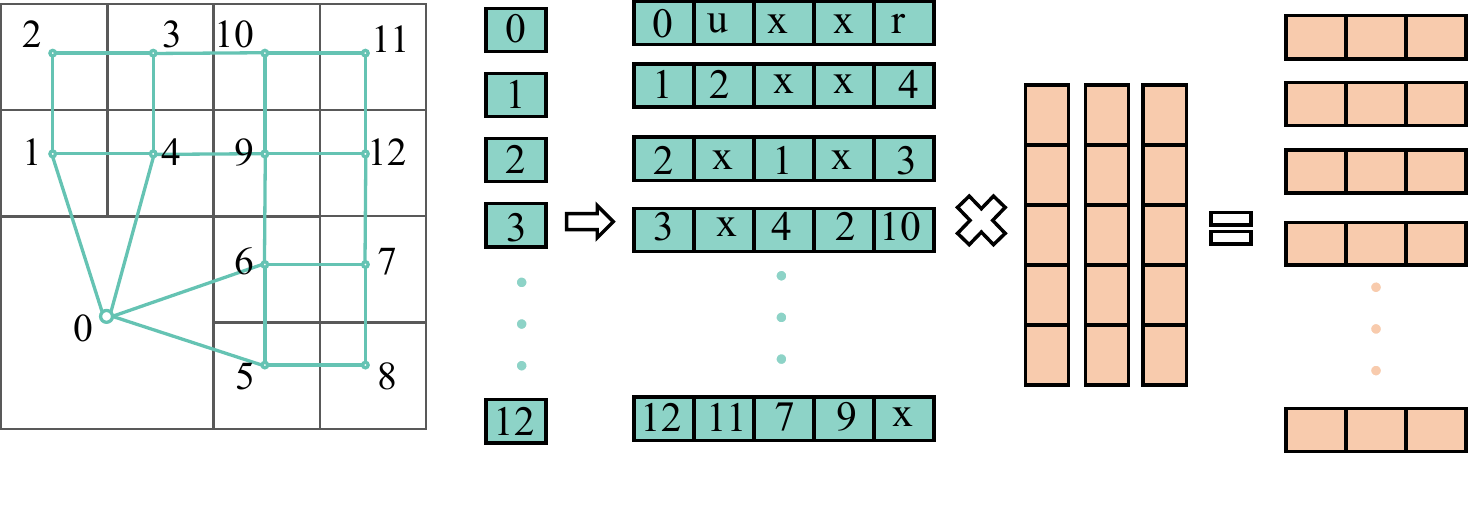}
    \put(8, -1){\large Dual octree}
    \put(32, -1){\large $F_{N \times C_i}$}
    \put(47, -1){\large $M_{N \times 7C_i}$}
    \put(70, -1){\large $W_{7C_i \times C_o}$}
    \put(89, -1){\large $F_{N \times C_o}$}
  \end{overpic}
  \caption{The computation of the proposed graph convolution with a 2D example.
  The numbers in the dual octree represent the indices of  node features.
  The matrix $M$ is constructed via \emph{torch.scatter}.
  The first column of $M$ stores the features of the centering nodes, and other columns store the features in up, down, left and right directions.
  In matrix $M$, x represents the node in the corresponding direction is missing and zeros are filled, $u$ in the first row of $M$ represents the sum of features at node 1 and node 4, $r$ represents the sum of features at node 5 and node 6.
  The updated features $F$ are computed via a matrix product of $M$ and a small weight matrix $W$.
  }
  \vspace{-5mm}
  \label{fig:gemm}
\end{figure}
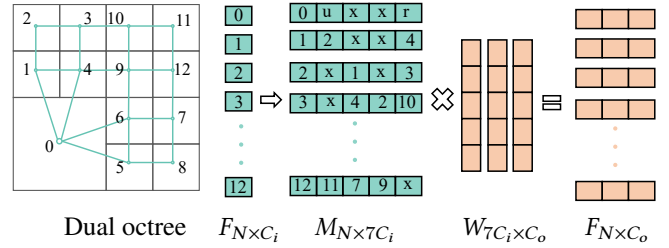

% implementation details
\paragraph{Implementation}
With our formulation in \cref{equ:conv}, the graph convolution can be implemented via a general matrix multiplication (GEMM), which executes efficiently on GPUs.
An example of graph convolution computation is shown in \cref{fig:gemm}.
We denote the node number of $\mathcal{V}^d$ by $N$, the channel of augmented node features by $C_i$, the channel of output node features by $C_o$, and the initial node features by $F_{N \times C_i}$.
According to the definition, the tensor shape of the weight matrix $W$ is $(C_i, 7, C_o)$.
For each node $v_i$, we first sum and gather the neighboring node features having the same directions with the efficient $torch.scatter$ operator provided by PyTorch, resulting in a tensor $M$ with shape $(N, C_i, 7)$.
After flattening the last two dimensions of $M$ and the first two dimensions of $W$, the updated node features can be calculated via a matrix product $F = M \times W$, which is executed efficiently by a highly optimized GEMM with a total complexity of  $\mathcal{O}(7N \times C_i\times C_o)$.
Thus, our graph convolution is much more efficient than previous point convolutions such as KPConv~\cite{Thomas2019} and NNConv~\cite{Simonovsky2017}.

\paragraph{Remark 1} Previous point-based graph convolutions are often applied to k-nearest-neighbor (KNN) graphs built from unorganized points. Our dual octree graph is different from KNN graphs. We empirically find that our scheme designed for dual octree graphs is more efficient and effective than prior KNN-based graph convolution schemes (see ablation studies in \cref{subsec:ablate}).

% further extensions.
\paragraph{Remark 2} 
The weight mapping function $W(\cdot)$ in \cref{equ:msg} can also be defined as $W(X_i, X_j, \Delta p_{ij})$, \ie conditioned on node features $X_i$ and $X_j$, apart from $\Delta p_{ij}$, as used by other works include DGCNN~\cite{Wang2019c}, graph attention convolutions~\cite{Velickovic2018}, and point cloud transformers~\cite{Guo2021,Zhao2021}.
Our dual graph convolution is compatible with this setup,  
and it is also possible to generate convolution weights conditioned on the voxel size of octree nodes and the varied density of input point clouds.
We leave the integration of these schemes for future work.

\subsubsection{Other graph operations} \label{subsec:sample}
% Graph downsampling and upsampling are used to generate hierarchical feature pyramids in the encoder and decoder.
% Skip connections are used to build connections between the encoder and decoder.

\paragraph{Graph Downsampling and Upsampling}
The dual octree graph contains leaf nodes with different resolutions.
The graph downsampling operation is performed by merging the \emph{finest} leaf nodes while keeping other octree nodes unchanged.
After downsampling, the octree depth decreases by 1, and accordingly the dual octree graph.
Specifically, we use a fully connected layer to merge the features of 8 sibling octree nodes with the finest resolution and generate the output feature vector of the corresponding parent node, which are concatenated with the unchanged coarse leaf node features.
Taking \cref{fig:dgraph} as an example, the dual octree graph changes from (b) to (c) after downsampling.
Compared with \cref{fig:dgraph}-(c), all the finest octree nodes in \cref{fig:dgraph}-(b) are merged and the largest depth of leaf nodes decreases by 1.
The graph upsampling operation is performed in reverse, simply by reverting the forward and backward computations of the downsampling operation.

\paragraph{Skip Connections}
We adapt the output-guided skip connection proposed by \cite{Wang2020}, which searches the corresponding and existing octree node in the encoder for each (generated) octree node in the decoder and adds the corresponding features from the encoder to the decoder. 
In our dual octree graph, each node $v_i$ is uniquely identified by $(x_i, y_i, z_i, d_i)$, which is used as a key in the searching process.
The output-guided skip connections are essential in building U-Net for the shape completion task, where the octrees in the encoder and decoder are different from each other.

\subsection{Neural MPU}  \label{subsec:mpu}

Our network extracts features for each octree node, while our goal is to obtain point-wise prediction.
To this end, we integrate the multi-level partition of unity (MPU)~\cite{Ohtake2003} into our pipeline.
MPU is a well-established method to blend locally defined functions into a global function by using locally-supported continuous weight functions that sum up to one everywhere on the domain.

Given a set of extracted node features $F^d=\{F_i\}$, we define an MLP $\Phi(x, F_i)$ for each octree node, which takes the node feature $F_i$ and local coordinates $x$ of a query point as input and outputs the local function value. On each octree node, we also define a local blending weight function $w_i(x)$, and associate a confidence value $c_i$ with $w_i(x)$, then the MPU can be formulated as
\begin{equation}
  F(x) = \frac{\sum_i c_i \cdot w_i(x) \cdot \Phi(x, F_i)} 
              {\sum_i c_i \cdot w_i(x)},
  \label{equ:mpu}
\end{equation}
where $F(x)$ is a global continuous and differentiable function,
and the weight function $w_i(x)$ is defined as
\begin{equation}
  w_i(x) = B(\frac{|x - o_i|} {r_i}), 
  \; \text{where} \;
  B(x) =
    \begin{cases}
      1 - |x| & \text{if}\; |x| < 1; \\
      0       & \text{otherwise}.
    \end{cases}
\end{equation}
Here $r_i$ and $o_i$ are the node cell size and the center of node $v_i$, respectively. $B(x)$ is a locally-supported and non-negative smooth function, and we set it as a linear B-Spline function. $c_i$ is the inverse of the volume of the corresponding octree node.  
To calculate the function value of a query point with \cref{equ:mpu}, we collect all $w_i(x)$ whose support region covers the point by following the efficient neighborhood searching scheme proposed in~\cite{Zhou2011}.

Due to the adaptiveness enabled by our dual octree graph, we can train a deep network outputting a feature volume with resolution up to $256^3$.
Therefore, we set $\Phi(x, F_i)$ as an MLP with a single hidden layer while retaining a good ability to encode geometric details of 3D shapes, which is contrary to previous methods using relatively deep MLPs~\cite{Park2019,Peng2020}.
In the evaluation stage, the node features $F^d=\{F_i\}$ are produced by a single forward pass of our network, and the prediction per point can be calculated simply by evaluating lightweight MLPs according to \cref{equ:mpu}.
Compared with deep MLPs, the computation cost is greatly reduced.
After predicting the field values, we use the marching cube method \cite{Lorensen1987} to extract the zero isosurfaces.

\subsection{Network and Loss functions} \label{subsec:details}

Given the basic network components in \cref{subsec:docnn}, we build a U-Net~\cite{Ronneberger2015} to predict the volumetric fields.
Also by removing the skip connections between the encoder and decoder of U-Net, we can get an autoencoder network structure.
The network can be trained in both supervised and unsupervised manners. 
Next, we detail the network and loss functions.

\begin{figure}[t]
  \centering
  \includegraphics[width=\linewidth]{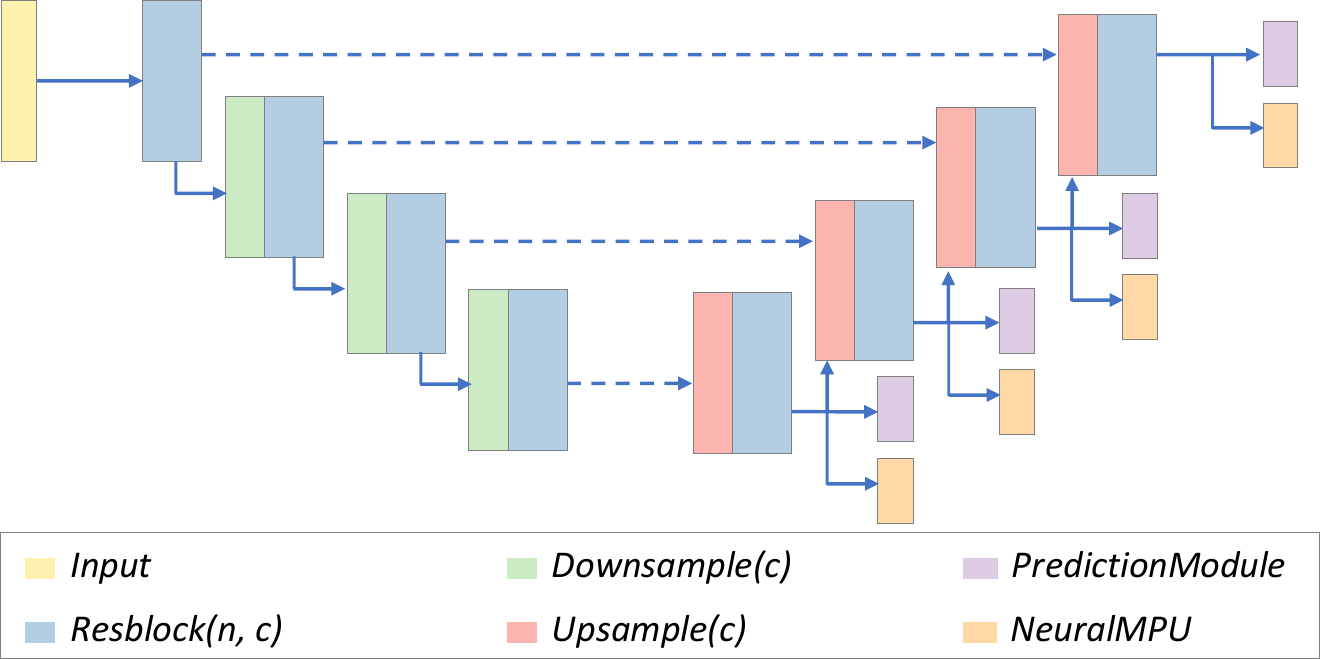}
  \caption{A U-Net built on our basic graph CNN operators. This network takes a dual octree graph with depth 6 as input, and predicts the octree splitting and corresponding volumetric fields.}
  \label{fig:unet}
\end{figure}

\paragraph{Network}
The overall network structure is shown in \cref{fig:unet}. 
We use $ResBlock(n, c)$ to represent a stack of $n$ ResNet blocks~\cite{He2016}.
Each ResNet block is constructed by concatenating two graph convolutions with a channel number of $c$.
$Downsample(c)$ and $Upsample(c)$ are the graph downsampling and upsampling operators based on shared FC layers with the input and output channel as $c$.
The extracted features at the end of each stage of the decoder are passed to a $PredictionModule$ to predict the octree splitting, which is made up of a shared MLP with 2 FC layers and 64 hidden units.
Meanwhile, the extracted features are also passed to $NeuralMPU$ (defined in \cref{subsec:mpu}) to predict the field values.
The U-Net shown in~\cref{fig:unet} takes a dual octree with depth 6 as input and output (resolution $64^3$), which is used for shape reconstruction in \cref{subsec:shapenet}. When dealing with octrees with larger depth ($>6$),
we add the required ResNet blocks and corresponding upsampling/downsampling operators accordingly based on the U-Net in~\cref{fig:unet}.
We set the maximum input/output octree depth as 7 (resolution $128^3$) in scene reconstruction in \cref{subsec:scene} and 8 (resolution $256^3$) in human body reconstruction in \cref{subsec:dfaust} for recovering the details of human hands and faces.
Note that the output resolution of our method can be different from the input resolution because the MPU can generate features for any
point in the 3D volume.
The channel number $c$ is set as 196 when the octree depth is 3, 128 when 4, then decreases by a factor of 2 when the depth increases.

\paragraph{Octree Loss}
When training the network to regress the ground-truth volumetric fields, we build an octree based on the ground-truth shapes as supervision.
We use a binary cross entropy loss to train the network to determine whether the node is empty or not, 
\begin{equation}
  \mathcal{L}_{octree} = \sum_d \frac{1}{N_d} \sum_{o \in \mathcal{O}_d} \operatorname{CrossEntropy}(o, o_{gt}),
  \label{equ:octree}
\end{equation}
where $d$ represents the depth of an octree layer, $N_d$ represents the total node number in the $d$-th layer, $O_d$ represents the predicted octree node status, and $o_{gt}$ is the corresponding ground-truth node status.

\paragraph{Regression Loss}
To regress the ground-truth field values, we use the following $L_2$ loss:
\begin{equation}
  \mathcal{L}_{regress} = 
    \sum_d \frac{1}{N_\mathcal{P}} \sum_{x \in \mathcal{P}} \left(
    \lambda_v \| F(x) - G(x) \|_2^2 + \| \nabla  F(x) -  \nabla G(x) \|_2^2 \right) ,
  \label{equ:regress}
\end{equation}
where $F(x)$ is the predicted volumetric fields, $G(x)$ is the ground-truth field values, and $\mathcal{P}$ is a set of query points.
The gradient $\nabla G(x)$ is computed via finite difference, and $\nabla F(x)$  is computed by auto-differentiation.
We set $\lambda_v$ as 200 to balance the two losses.
The regression loss is also imposed on every layer of the octree, which we empirically find that it can slightly improve the final performance.

\paragraph{Gradient Loss}
In surface reconstruction, the ground-truth field values are not known. We use the following gradient loss inspired by Posision surface reconstruction~\cite{Kazhdan2006}:
\begin{equation}
  \begin{split}
  \mathcal{L}_{grad} = \sum_d \Big\{ 
    & \frac{1}{N_\mathcal{S}}  \sum_{x \in \mathcal{S}} \left(
       \lambda_v \| F(x) \|_2^2 + \| \nabla  F(x) -  \mathcal{N}(x) \|_2^2 \right) + \\
    &  \frac{1}{N_\mathcal{Q}} \sum_{x \in \mathcal{Q}} 
       \lambda_g \| \nabla F(x) \|_2^2 \Big\} ,
  \label{equ:grad}
  \end{split}
\end{equation}
where $\mathcal{S}$ is a set of points from the input point cloud, $\mathcal{Q}$ is a set of points sampled uniformly from the 3D volume and $\mathcal{Q} \bigcap \mathcal{S} = \emptyset$,  $\mathcal{N}(x)$ returns the normal of a point $x$ in $\mathcal{S}$, $\lambda_v$ and $\lambda_g$ is set to 200 and 0.1 respectively.
$\mathcal{L}_{grad}$ encourages $F(x) = 0,  \nabla F(x) = \mathcal{N}(x), \forall x \in \mathcal{S}$ and $ \nabla F(x) = 0, \forall x \in \mathcal{Q}$,
resulting in an occupancy field that approximates the input points and normals, and roughly equals $0.5$ outside the shape and $-0.5$ inside.
Here, notice that the third term is different from the Eikonal loss used by IGR~\cite{Gropp2020}, as the predicted volumeric field is an occupancy field, not an SDF field. We find that our occupancy field based loss function is not sensitive to network weight initialization, thus initializing our network randomly, while the SDF-field based loss design by IGR~\cite{Gropp2020} needs a special geometric initialization~\cite{Atzmon2020}.

\section{Experiments and Discussions} \label{sec:result}

We validate the efficiency and efficacy of our network on a series of 3D reconstruction tasks, including shape or scene reconstruction from noisy point clouds, unsupervised surface reconstruction, and autoencoders, and demonstrate its superiority to other state-of-the-art methods. All our experiments are conducted on 4 Nvidia V100 GPUs with 32GB memory.

\subsection{Shape Reconstruction from Point Clouds} \label{subsec:shapenet}

In this section, we train a network to reconstruct the volumetric representation of 3D shapes from noisy and sparse input point clouds.
The ground-truth volumetric fields are needed during the training stage. Therefore, this task can also be regarded as a kind of shape super-resolution.

\paragraph{Dataset}
We use man-made CAD shapes of 13 categories from ShapeNet to train the network and evaluate the performance following ConvONet~\cite{Peng2020}, with the same data splitting.
The original meshes of ShapeNet contain artifacts including flipped and overlapped triangles, self-intersections, and non-manifold geometry, we preprocess them to watertight meshes using the approach proposed in~\cite{Xu2014} first, then normalize the meshes into a unit bounding box defined in $[-0.5, 0.5]^3$ and compute the ground-truth SDF with the fast marching algorithm.
To generate the input noisy point clouds, we randomly sample 3000 points for each mesh and add Gaussian noise with a standard deviation of 0.005, following the same sampling strategy as~\cite{Mescheder2019,Peng2020}.
For sampling the ground-truth SDF values to train the network, we first build octrees with depth 6 on watertight meshes, randomly sample 4 query points in each leaf node of the octree, and attach the corresponding SDF values to these points.
This strategy resembles the sampling methods used in~\cite{Park2019,Liu2021b}, with which we can sample more points near the surface.
These SDF samples are computed in the preprocessing step, and a subset is randomly drawn to evaluate the loss function in each training iteration.\looseness=-1

\begin{table}[t]
  \tablestyle{5pt}{1.1}
  % \centering
  \caption{Quantitative evaluation on the ShapeNet dataset. The numbers outside and inside the parentheses are the results on the testing dataset of 13 categories from ShapeNet and on the 5 unseen categories respectively. The Chamfer distance (CD) is multiplied by a factor of 100 for better display.
  }
  \begin{tabular}{ccccc}
    \toprule
    Network      & CD$\downarrow$      & NC$\uparrow$         & IoU$\uparrow$           & F-Score$\uparrow$    \\ % & Time$\downarrow$  \\
   \midrule
    ConvONet     & 0.438 (0.487)       & 0.932 (0.951)        & 0.852 (0.845)           & 0.925 (0.912)        \\ % & - \\
    DeepMLS      & 0.271 (0.302)       & 0.945 (0.957)        & 0.891 (0.894)           & 0.988 (0.985)        \\ % & - \\
    Ours         & \tbf{0.267 (0.286)} & \tbf{0.953 (0.967)}  & \tbf{0.904 (0.918)}     & \tbf{0.991 (0.990)}  \\ % & - \\
    \bottomrule
  \end{tabular}
  \label{tab:shapenet}
  \vspace{-4mm}
\end{table}

\paragraph{Evaluation Metrics}
The output meshes are extracted with Marching Cubes~\cite{Lorensen1987} on a volumetric grid of $128^3$.
Following ConvONet~\cite{Peng2020}, we use the $L_1$ Chamfer distance (CD), normal consistency (NC), volumetric IoU, and F-Score as the evaluation metrics to measure the quality of the predicted meshes, among which the first two metrics measure the surface quality of the extracted meshes compared with ground-truth meshes and the last two metrics measure the volumetric quality of the output.

We denote the predicted mesh and ground-truth mesh by $M_P$ and  $M_G$, on which we randomly sample a set of $N$ ($10k$) points $P = \{x_i\}_{i=1}^N$ and  $G = \{y_i\}_{i=1}^N$, respectively.
Define $\mathcal{P}_{A}(x) = \arg \min_{y \in A} \| x - y\| $, which finds the closest point of $x$ from a point set $A$.
The $L_1$ Chamfer distance is defined as
\begin{equation}
  CD = \frac{1}{N} \sum_i \| x_i - \mathcal{P}_{G}(x_i) \| +
  \frac{1}{N} \sum_i \| y_i - \mathcal{P}_{P}(y_i) \|.
  \label{equ:chamfer}
\end{equation}
We define $\mathcal{N}(x)$ as an operator that returns the corresponding normal of an input point, then the normal consistency is defined as
\begin{equation}
  \begin{aligned}
    NC = \frac{1}{N} \sum_i |\mathcal{N}(x_i) \cdot
    \mathcal{N}(\mathcal{P}_{G}(x_i))| +
    \frac{1}{N} \sum_i |\mathcal{N}(y_i) \cdot
    \mathcal{N}(\mathcal{P}_{P}(y_i))|.
  \end{aligned}
  \label{equ:normal}
\end{equation}
The volumetric IoU is defined as the volumetric intersection over union between  $M_P$ and  $M_G$,
% \begin{equation*}
%   IoU = \frac{| M_P \bigcap M_G |}{| M_P \bigcup M_G |}
% \end{equation*}
and we sample 100k points in the 3D volume to approximate the results.
The F-Score is defined as the harmonic mean between the precision and the recall of points that lie within a certain distance between $M_G$ and $M_P$.

\paragraph{Implementation Details}
We use Adam~\cite{Kingma2014a} as the optimizer and train the network in 300 epochs with a batch size of 16. The initial learning rate is set as $10^{-3}$ and decreases to $10^{-5}$ linearly throughout the training process.
The loss function is simply a summation of the octree loss (\cref{equ:octree}) and the regression loss  (\cref{equ:regress}):
$\mathcal{L} = \mathcal{L}_{octree} + \mathcal{L}_{regress}$.
In each iteration, we randomly sample 5000 points from the ground-truth surface and 5000 points from the pre-computed SDF samples to evaluate the loss.

\begin{figure*}
  \centering
  \begin{overpic}[width=0.98\linewidth]{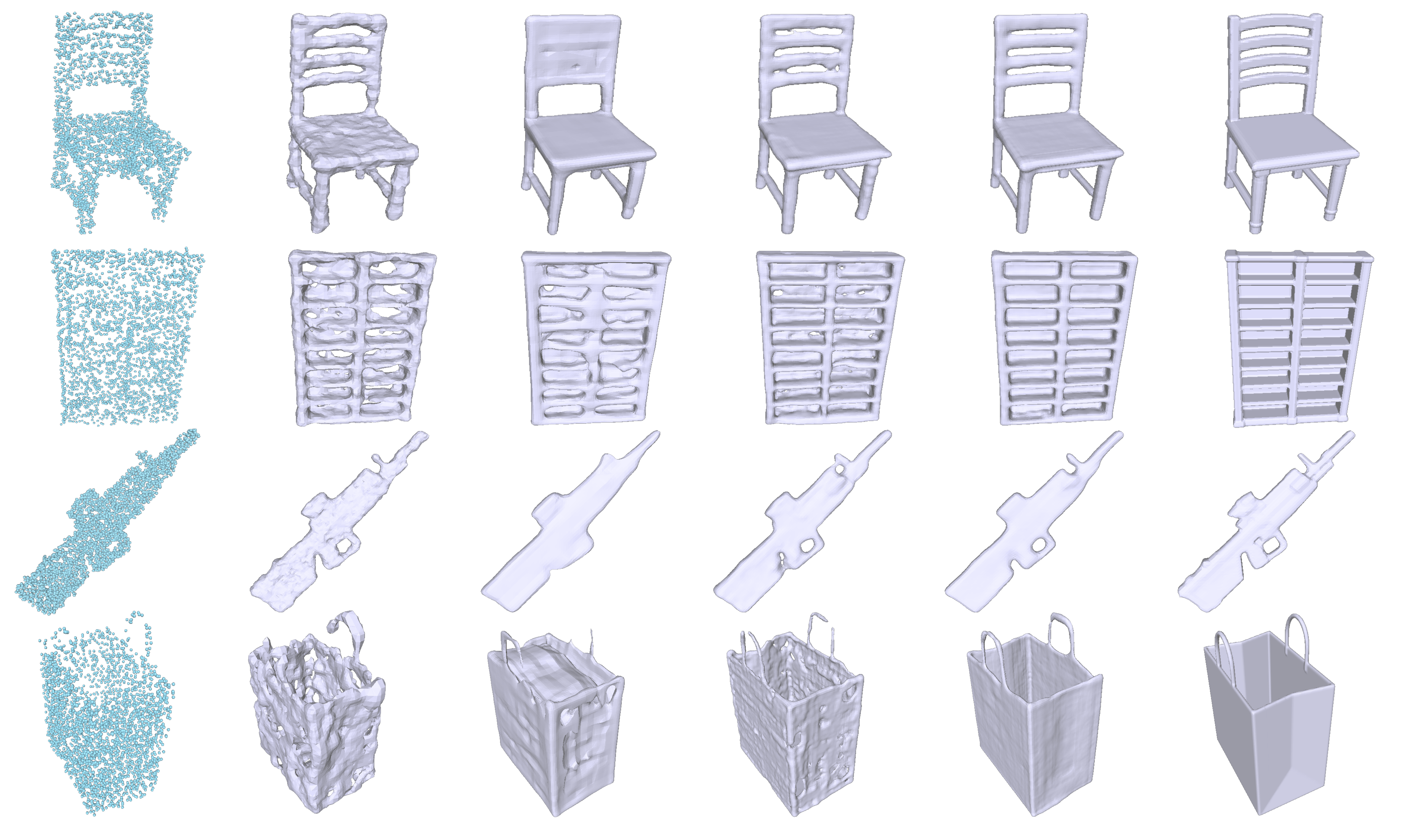}
    \put(2,  -0.5){\small (a) Point cloud}
    \put(22, -0.5){\small (b) PSR}
    \put(36, -0.5){\small (c) ConvONet}
    \put(53, -0.5){\small (e) DeepMLS}
    \put(72, -0.5){\small (d) Ours}
    \put(86, -0.5){\small (e) Ground-truth}
  \end{overpic}
  % \vspace{4pt}
  \caption{Shape-level reconstruction from noisy point clouds.
  The first three rows show the results from the testing dataset, and the last row shows the results of an unseen shape.}
  \label{fig:shapenet}
\end{figure*}

\paragraph{Results}
We conduct a comparison with two recent learning-based methods that achieved top performance on this benchmark, including ConvONet~\cite{Peng2020},  Deep Moving Least Squares (DeepMLS)~\cite{Liu2021b} and the classical Poisson surface reconstruction (PSR)~\cite{Kazhdan2006}.
Our network has 2.2M parameters, which is between DeepMLS (4.6M) and ConvONet (1.1M).
The numerical results are shown in \cref{tab:shapenet}, our method clearly achieves the best performance on all metrics.
The visual comparisons are shown in \cref{fig:shapenet}.
Since PSR needs point normals as input, we assign ground-truth normals to each point.
It can be seen that the results of PSR are noisy and contain holes in the reconstructed shapes, even with ground-truth normals as additional input.
The results of ConvONet are over-smoothed and lack fine geometric details.
DeepMLS combines the octree-based U-Net~\cite{Wang2020} and moving least squares~\cite{Oztireli2009}, and has higher expressiveness to reproduce the geometric details than ConvONet. 
However, DeepMLS also introduces high-frequency errors in the results, as shown in the back of the chair and the cells of the cabinet in \cref{fig:shapenet}.
Moreover, DeepMLS constrains the output volumetric fields near the surface, the predicted representation of shapes is localized and the field values are undefined far away from the surface.
Our network reconstructs the whole volumetric field made up of adaptive voxels: the low-resolution voxels help our network to fit the low-frequency part of the volumetric field, and the high-resolution voxels enable our network to reconstruct the fine geometric details.
Therefore, our results are much more faithful to the ground-truth meshes visually and numerically.

\paragraph{Generalization Ability}
To test the generalization ability of our network, we forward the networks on 5 unseen categories of ShapeNet, including bathtub, bag, bed, bottle, and pillow.
We calculate the evaluation metrics and summarize the results in parentheses of \cref{tab:shapenet}.
Our network achieves the highest performance on all metrics again. 
And the evaluation metrics do not change too much on these 5 unseen categories compared to the results on the testing dataset, which verifies that our network has strong generalization ability.
The visual comparison is shown in the last row of \cref{fig:shapenet}, our network produces the best result and is significantly better than the others. \looseness=-1

\begin{table}[t]
  \tablestyle{10pt}{1.1}
  \caption{Ablation study on the necessity of incorporating multiscale voxels in the convolution. O-CNN and \textsc{SingleScale} involve voxels of single scale in the convolution. \textsc{Fine$\rightarrow$Coarse} adds coarse voxels  for fine-scale voxels when doing convolution. \textsc{FullGraph} operates on the full dual octree graph. }
  \begin{tabular}{ccccc}
    \toprule
    Network      & CD$\downarrow$    & NC$\uparrow$    & IoU$\uparrow$  & F-Score$\uparrow$    \\
   \midrule
   \textsc{O-CNN}                        & 0.283          & 0.949          & 0.891       & 0.987            \\
   \textsc{SingleScale}                  & 0.290          & 0.948          & 0.886       & 0.987            \\
   \textsc{Fine$\rightarrow$Coarse}      & 0.269          & 0.951          & 0.901       & 0.989            \\
   \textsc{FullGraph}                    & \tbf{0.267}    & \tbf{0.953}    & \tbf{0.904} & \tbf{0.991}      \\
  \bottomrule
  \end{tabular}
  \vspace{-2mm}
  \label{tab:ablation1}
\end{table}

% \begin{figure}[t]
%   \centering
%   \begin{overpic}[width=0.98\linewidth]{sdf}
%     \put(5, 33.5){\small (a) Predicted SDF}
%     \put(44, 33.5){\small (b) SDF}
%     \put(72, 33.5){\small (c) Union}
%     \put(10, -1){\small (d) -0.07}
%     \put(43, -1){\small (e) 0.1}
%     \put(76, -1){\small (f) 0.2}
%   \end{overpic}
%   % \vspace{4pt}
%   \caption{More applications of the output of our network.
%   (a) is a slice of the SDF predicted by our network.
%   (b) is a slice of the SDF of SIGGRAPH.
%   (c) we can easily get the union of (a) and (b) by applying a \emph{min} operator to them.
%   The meshes from (d) to (f) are the different isosurfaces of the predicted SDF (a).}
%   \label{fig:sdf}
% \end{figure}

% \paragraph{More Applications of SDFs}
% Our network predicts the SDFs in the whole volume, which enables many shape modeling operations.
% In \cref{fig:sdf}-(a), we show a slice of the predicted SDF by our network.
% With SDFs, the boolean operations between two shapes can be implemented efficiently, for example, we can get the union of two shapes by using an element-wise \emph{min}, as shown in \cref{fig:sdf}-(c).
% Moreover, we can also conveniently dilate or shrink the shapes by extracting different isosurfaces of the predicted SDF, as shown in the second row of \cref{fig:sdf}.

\begin{figure*}
  \centering
  \begin{overpic}[width=0.96\linewidth]{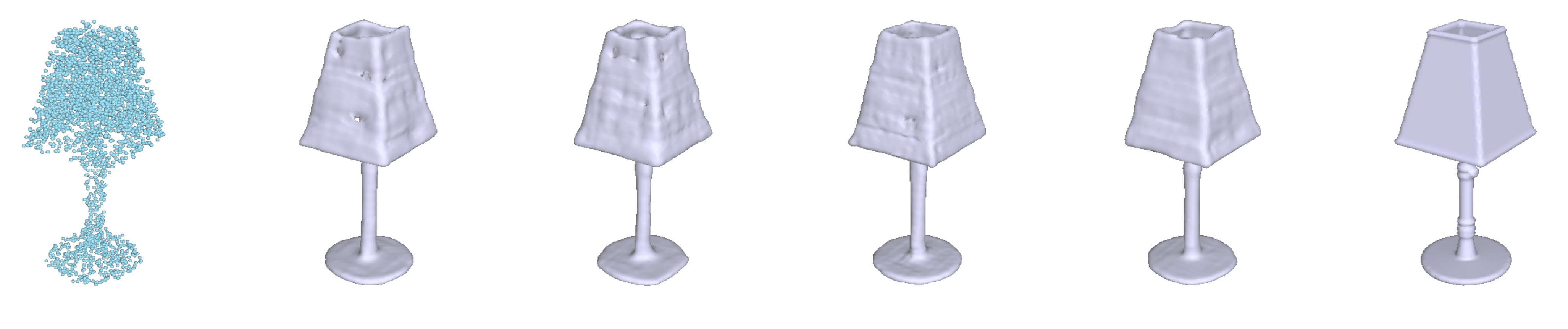}
    \put(2,  -0.5){\small (a) Point cloud}
    \put(18, -0.5){\small (b) \textsc{SingleScale}}
    \put(38, -0.5){\small (c) KPConv}
    \put(55, -0.5){\small (e) EdgeConv}
    \put(74, -0.5){\small (d) Ours}
    \put(88, -0.5){\small (e) Ground-truth}
  \end{overpic}
  % \vspace{4pt}
  \caption{Visual comparison for different graph convolution designs.}
  \label{fig:ablation}
\end{figure*}
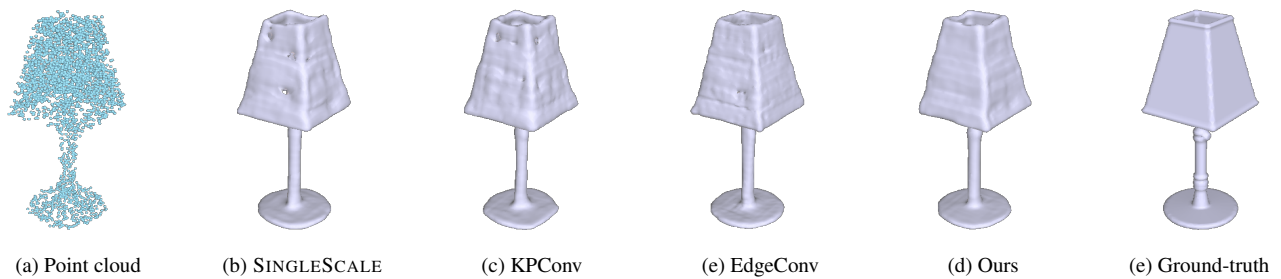

\subsection{Ablation Studies and Discussions} \label{subsec:ablate}
In this section, we study several key design choices and discuss the efficiency of our method on top of the experiment in~\cref{subsec:shapenet}.

\paragraph{Incorporate Multiscale Features in Convolutions}
One of the key properties of our graph convolution is to keep all voxels covering the whole volume as active and involve multiscale voxel features in each convolution.
To justify the necessity of this strategy, we design the following baselines.
\begin{itemize}[leftmargin=10pt]
  \item[-] O-CNN: Use octree-based sparse convolution (O-CNN)~\cite{Wang2017} in the network. By combining O-CNN with our neural MPU, the network can also produce continuous volumetric fields. In this setting, only octree nodes with the same scale are involved in each convolution. 
  \item[-] \textsc{SingleScale}: Remove all cross-scale edges in the dual octree graph, which makes the graph convolution resemble O-CNN. The only difference is that the kernel shape of O-CNN is a cube with 27 neighborhoods, whereas the kernel shape of the graph convolution in this setting is a 3D cross with 7 neighborhoods. For a fair comparison, we reduce the channel size of O-CNN accordingly to ensure that the number of trainable parameters is similar.
  \item[-] \textsc{Fine$\rightarrow$Coarse}: Based on \textsc{SingleScale}, add graph edges from fine-scale nodes to coarse-scale nodes so that fine octree nodes can exploit coarse-scale features during convolution.
\end{itemize}
We denote \textsc{FullGraph} as our default implementation that applies the graph convolution in the full dual octree.
We use the same network settings and training configurations as in \cref{subsec:shapenet} and conduct experiments on ShapeNet.
The results are summarized in \cref{tab:ablation1}.
Based on the experiment results, we make the following observations:
\begin{itemize}[leftmargin=10pt]
  \item[-] With a similar amount of trainable parameters, \textsc{SingleScale} is only slightly worse than O-CNN, even though the kernel size of \textsc{SingleScale} is smaller than O-CNN.
  \item[-] By simply adding the edges from fine-scale to coarse-scale nodes, \textsc{Fine$\rightarrow$Coarse} is much better than \textsc{SingleScale} and O-CNN.
  \item[-] By adding all edges to the dual octree graph, the performance of \textsc{FullGraph} improves further compared with \textsc{Fine$\rightarrow$Coarse}.
\end{itemize}
The visual comparison is shown in \cref{fig:ablation}: the result of \textsc{SingleScale} contains unwanted holes compared with \textsc{FullGraph}.
Given these observations, we draw the following conclusions:
\begin{itemize}[leftmargin=10pt]
  \item[-] It is necessary to involve multiscale node features when performing convolutions. The coarse-scale features can help increase the receptive field and eliminate high-frequency errors.
  \item[-] For cross-scale edges, the edges from coarse-scale nodes to fine-scale nodes are more critical for the final performance.  We suspect that the fine-scale nodes have only a small volume coverage, thus being only slightly helpful for coarse-scale nodes.
\end{itemize}

\begin{table}
  \tablestyle{6pt}{1.1}
  \caption{Performance comparisons among our graph convolution, KPConv and EdgeConv.}
  % \centering
  \begin{tabular}{cccccc}
  \toprule
    Network               & CD$\downarrow$  & NC$\uparrow$    & IoU$\uparrow$    & F-Score$\uparrow$    & Train. Time$\downarrow$\\
   \midrule
  %  \textsc{NNConv}      & -               & -            & -           & -             &   \\
    KPConv         & 0.291           & 0.944        & 0.889       & 0.984         &  \SI{89.1}{\hour}  \\
    EdgeConv       & 0.289           & 0.945        & 0.891       & 0.982         &  \SI{96.5}{\hour}  \\
    Ours           & \tbf{0.267}     & \tbf{0.953}  & \tbf{0.904} & \tbf{0.991}   & \tbf{\SI{17.4}{\hour}} \\
  \bottomrule
  \end{tabular}
  \vspace{-2mm}
  \label{tab:ablation2}
\end{table}

\begin{figure*}
  \centering
  \begin{overpic}[width=\linewidth]{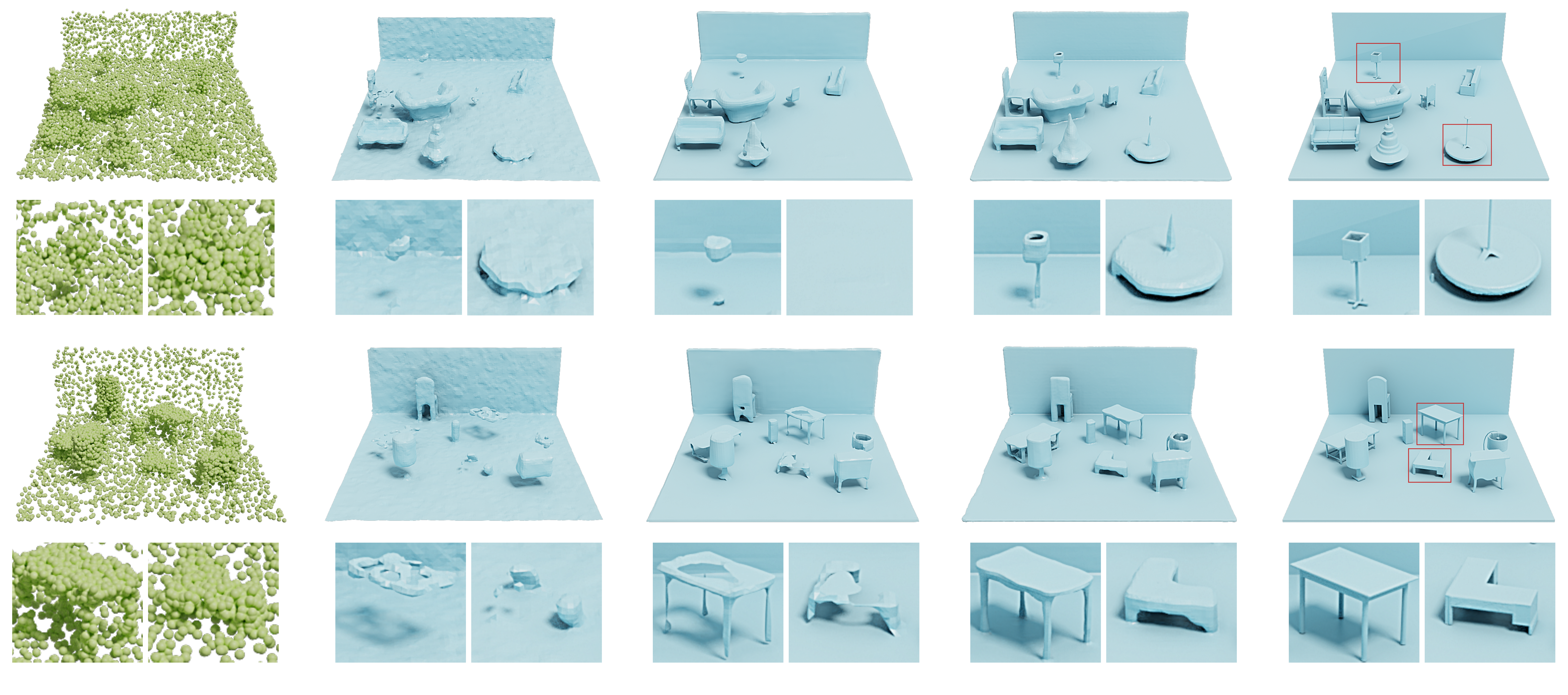}
    \put(5,  0){\small (a) Point cloud}
    \put(28, 0){\small (b) PSR}
    \put(46, 0){\small (c) ConvONet}
    \put(68, 0){\small (d) Ours}
    \put(86, 0){\small (e) Ground-truth}
  \end{overpic}
  \caption{Scene-level reconstruction from noisy point clouds.
    % Our results in column (d) are more faithful to the ground-truth meshes.
    The results of PSR in column (b) are noisy and incomplete, and the results of ConvONet in column (c) lack fine geometric details and even miss reconstructing some objects in the scene, as highlighted in the zoom-in images.}
  \label{fig:scene}
\end{figure*}
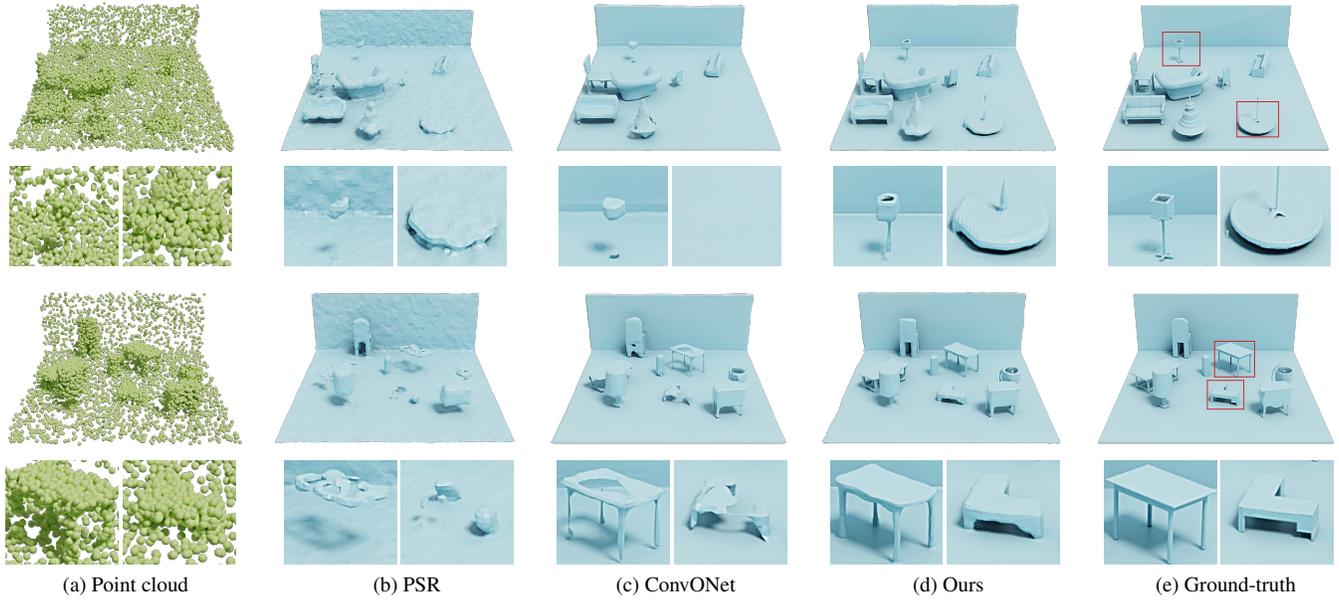

\paragraph{Comparison with Other Point Convolutions}
We compare our graph convolution tailored for dual octree graphs with previous widely-used point cloud convolutions, including EdgeConv~\cite{Wang2019c} and KPConv~\cite{Thomas2019},  to highlight the effectiveness and efficiency.
In DGCNN~\cite{Wang2019c}, EdgeConv is originally applied to a dynamic  \textsc{KNN} graph built in the feature space, we instead apply EdgeConv to our dual octree graph, otherwise, the network will be out-of-memory in runtime. We set the edge function of EdgeConv as an MLP with 2 hidden layers and 64 hidden units.
For KPConv, we use its rigid version, which is similar to SplineConv~\cite{Fey2018}.
We set the radius of KPConv to the largest edge length of dual octree graphs. 
We also tried PointCNN~\cite{Li2018} and NNConv~\cite{Simonovsky2017}, the training of the corresponding networks failed due to being out-of-memory.

We use the highly optimized implementations from the PyTorch Geometric library~\cite{Fey2019} to test these alterative graph convolution schemes. 
We use the same training configurations in \cref{subsec:shapenet}.
The evaluation metrics are summarized in \cref{tab:ablation2}.
We also measure the total training time on 4 V100 GPUs, which are shown in the last column of \cref{tab:ablation2}.
Our network not only achieves the best performance but also runs 5.1 times and 5.5 times faster than KPConv and EdgeConv respectively, justifying the benefits of taking into account the specific structure of the dual octree graph when designing the graph convolution.
The visual comparisons shown in \cref{fig:ablation} also verify that with our graph convolution the network produces better results.

%\vspace{-12pt}
\paragraph{Benefits of Neural MPU}
The neural MPU in~\cref{equ:mpu} is used to guarantee the continuity of the resulting 3D field by interpolating features with different resolutions produced by our graph network.
The trilinear interpolation used in ConvONet~\cite{Peng2020} can only deal with features defined on a 3D uniform grid.
ACORN~\cite{Martel2021} can deal with adaptive features, whereas the resulting 3D fields are not continuous.
The effect of our neural MPU can be verified by the extracted smooth iso-surfaces shown in \cref{fig:shapenet}.

%\vspace{-12pt}
\paragraph{Gradient Term in $\mathcal{L}_{regress}$}
In $\mathcal{L}_{regress}$, we not only regress the ground-truth volumetric fields, but also add a gradient loss term following DeepMLS~\cite{Liu2021b}.
We observe that the gradient loss is essential to improve the quality of the reconstructed 3D field.
Without it, the performance of our method drops slightly (CD: 0.269, NC: 0.949, IoU: 0.901, F-Score: 0.990), but it still outperforms DeepMLS and ConvONet in~\cref{tab:shapenet}.
\begin{table}
  \tablestyle{4pt}{1.1}
  \caption{Time and GPU memory cost of our network, measured on a NVidia V100 GPU.
  The network takes a batch of 16 shapes for training and one single shape for inference, excluding marching cubes.}
  \begin{tabular}{ccccc}
    \toprule
    N            & 32           & 64            & 128           & 256           \\ 
   \midrule
   Training     & 6.1GB/174ms & 10.7GB/271ms & 16.8GB/489ms & 27.2GB/924ms \\ 
   Inference    & 0.4GB/41ms  & 0.7GB/55ms   & 1.1GB/91ms  & 1.5GB/153ms \\ 
   \bottomrule
  \end{tabular}
   \vspace{-3mm}
  \label{tab:time}
\end{table}

%\vspace{-12pt}
\paragraph{Time and Memory Efficiency}
The number of octree nodes increases quadratically with respect to volumetric resolution $N$, so our network has $\mathcal{O}(N^2)$ memory and computational cost.
We record the average memory and computational cost of our network  with a batch of 16 shapes with a resolution of 64 for training and with one single shape for inference (excluding marching cubes) on a V100 GPU, and report the results in~\cref{tab:time}.
As a reference, the memory footprints of the networks in~\cref{tab:shapenet} are \SI{9.4}{GB} for DeepMLS and \SI{8.7}{GB} for O-CNN with the same settings.
ConvONet adopts a vanilla volumetric CNN, takes 7.5G memory with a resolution of 32, and runs out of memory with a resolution of 64 on the ShapeNet dataset.
In summary, our network is as efficient as other octree-based networks, and much more efficient than volumetric CNNs.

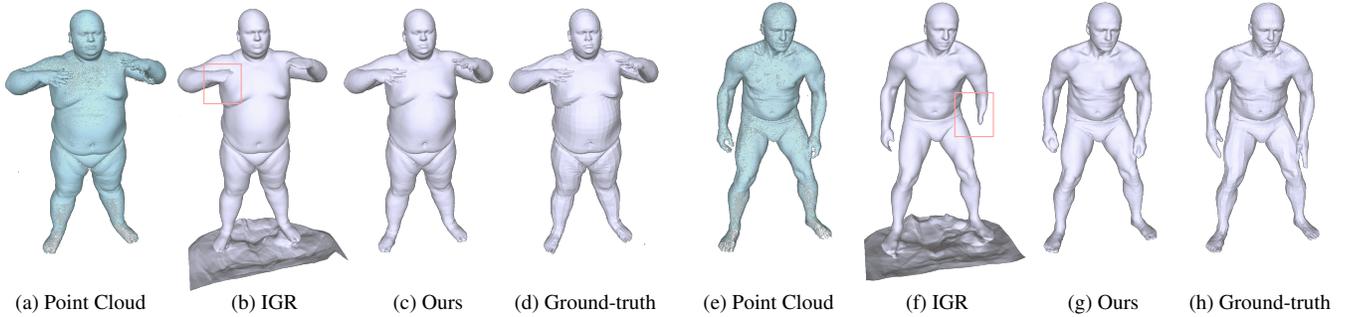
\begin{figure*}[t]
  \centering
  \begin{overpic}[width=\linewidth]{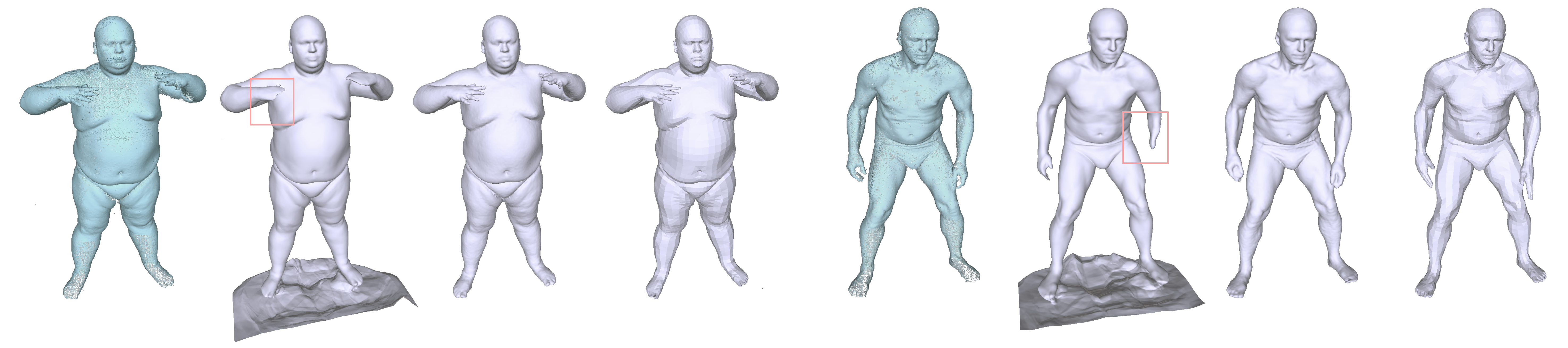}
    \put( 2, -1){\small (a) Point Cloud}
    \put(18, -1){\small (b) IGR}
    \put(30, -1){\small (c) Ours}
    \put(39, -1){\small (d) Ground-truth}
    \put(53, -1){\small (e) Point Cloud}
    \put(68, -1){\small (f) IGR}
    \put(80, -1){\small (g) Ours}
    \put(89, -1){\small (h) Ground-truth}
  \end{overpic}
  \caption{Visual results of surface reconstruction on D-Faust.
  IGR cannot reconstruct the hands of the body shapes, as highlighted by the red box. IGR produces fake surface sheets on the feet.}
  \label{fig:dfaust}
\end{figure*}

\subsection{Scene Reconstruction from Point Clouds} \label{subsec:scene}

In this section, we conduct experiments on scene-level datasets to verify the scalability of our method.
The network takes a noisy point cloud sampled from scene-level data as input and regresses the corresponding volumetric occupancy fields.

\paragraph{Dataset}
We use the synthetic scene dataset provided by ConvONet~\cite{Peng2020}.
The dataset contains $5k$ scenes with randomly selected objects from 5 categories of ShapeNet, including chair, sofa, lamp, cabinet, and table.
Following ConvONet, we uniformly sample $10k$ points from each scene, add Gaussian noise with a standard deviation of 0.05 as input, and let the network regress the ground-truth occupancy values.
We also use the same data splitting as ConvONet.
We set the octree depth as 7, and the equivalent maximum voxel resolution is $128^3$.
% After training the network on the synthetic dataset, we also test the generalization ability on real data ScanNet~\cite{Dai2017a}, which contains 1513 indoor scenes captured by a real 3D scanner.

\paragraph{Training Details}
We use Adam~\cite{Kingma2014a} as the optimizer with a batch size of 16. The initial learning rate is set as $10^{-3}$ and decreases to $10^{-5}$ linearly throughout the training process.
We trained the network in 900 epochs, with the similar training iteration number as ConvONet.
The loss function is defined as $\mathcal{L} = \mathcal{L}_{octree} + \mathcal{L}_{regress}$.
In each iteration, we randomly sample 10000 points from the ground-truth surfaces and 10000 points from ground-truth occupancy values in the 3D volume to evaluate the loss.

\paragraph{Results}
We compare our method with ConvONet~\cite{Peng2020}, ONet~\cite{Mescheder2019} and PSR~\cite{Kazhdan2006}.
We calculate the metrics defined in \cref{subsec:shapenet}, and the results are shown in \cref{tab:scene}.
For ConvONet, we use the code released by the authors to generate the results. 
For ONet, we directly reuse the results provided by \cite{Peng2020}.
Our method is consistently better than the other methods.
ConvONet uses dense voxels with resolution $64^3$ in the network.
However, due to the constraints of computation and memory caused by dense voxels, the CNN layers operating on higher resolution features are shallow and have low expressiveness.
Our network directly outputs adaptive volumetric grids with the finest resolution up to $128^3$, and our network has a stack of CNN blocks even on high-resolution features, so our network can reconstruct fine details faithfully.
% We attribute our performance improvement over ConvOnet to the efficiency and scalability enabled by our network based on dual octree graph convolutions.
We also record the running time of network forwarding and field evaluation on $128^3$ uniformly sampled query points in the testing stage, excluding the time of marching cubes and hard disk I/O.
\cref{tab:scene} shows our network is 2.7 times faster than ConvONet in performing this task.

The visual results are shown in \cref{fig:scene}, it is obvious that our results are more faithful to the ground-truth meshes.
The results of PSR are noisy and incomplete due to the lack of priors from the dataset.
ConvONet cannot reproduce fine details and may also fail to reconstruct some objects in the scene, as highlighted in \cref{fig:scene}.

\begin{table}
  \tablestyle{8pt}{1.1}
  \caption{Quantitative evaluation on the scene-level dataset. The results of ONet are provided by ConvONet. }
  \begin{tabular}{cccccc}
    \toprule
    Network      & CD$\downarrow$     & NC$\uparrow$    & IoU$\uparrow$  & F-Score$\uparrow$  & Time$\downarrow$  \\
   \midrule
    % Poisson$^*$  & 0.690              & 0.890           & -               & -            & -  \\
    ONet         & 2.100              & 0.783           & 0.475           & -            & -  \\
    ConvONet     & 0.419              & 0.914           & 0.849           & 0.964        & 0.857s  \\
    Ours         & \tbf{0.388}        & \tbf{0.923}     & \tbf{0.857}     & \tbf{0.969}  & \tbf{0.314s}  \\
    \bottomrule
  \end{tabular}
  % \vspace{-5mm}
  \label{tab:scene}
\end{table}

\begin{figure*}
  \centering
  \begin{overpic}[width=\linewidth]{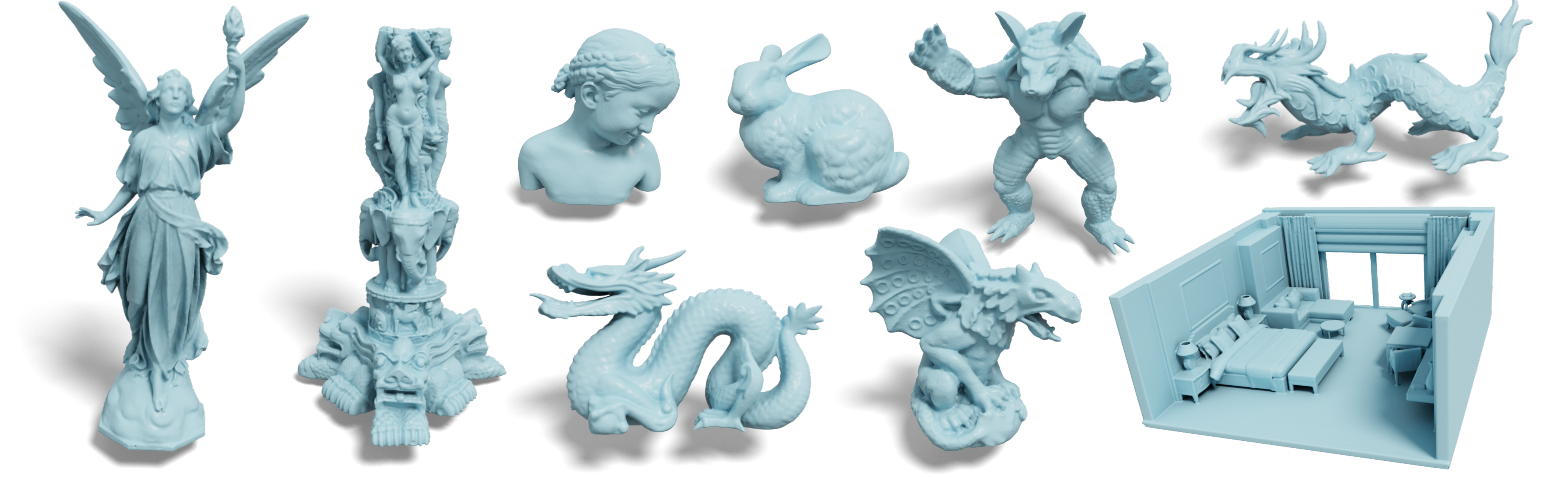}
  \end{overpic}
  \vspace{-6mm}
  \caption{Surface reconstruction on shapes with rich geometric details. The network trained on D-Faust directly takes a point cloud as input and predicts the corresponding volumetric field by a single forward pass. }
  \label{fig:surface}
  \vspace{-2mm}
\end{figure*}

\subsection{Unsupervised Surface Reconstruction}  \label{subsec:dfaust}

In this section, we focus on surface reconstruction from raw point clouds.
Surface reconstruction is a fundamental problem in computer graphics and has drawn much attention over the past few decades~\cite{Berger2017}.
In surface reconstruction, ground-truth volumetric field values are unknown or impossible to compute in advance, 
therefore the network is trained without the supervision of ground-truth volumetric field values.

\paragraph{Dataset}
We use the D-Faust dataset~\cite{Bogo2017} following IGR~\cite{Gropp2020}.
D-Faust contains real scans of human bodies in different poses, in which $6k$ scans are used for training and $2k$ scans for testing.
We use the raw point clouds with normals as input and set the octree depth to 8, with an equivalent volumetric resolution of $256^3$.
The point clouds are incomplete and noisy, caused by occlusion in the scanning process and limited precision of scanners.

\paragraph{Training details}
The training settings are the same as \cref{subsec:scene}, except that we set the maximal training epoch to 600.
Since there are no ground-truth field values, we trained the network with $\mathcal{L}_{grad}$ in \cref{equ:grad}.
The octrees in the encoder and decoder are the same, therefore we do not use the octree loss $\mathcal{L}_{octree}$.
In each training iteration, we randomly sample 10k points from the point cloud and the 3D volume for each shape to evaluate the loss function.

\begin{table}
  \tablestyle{6pt}{1.1}
  \caption{Quantitative evaluation on D-Faust. 
  CD(pred, gt) is the one-side chamfer distance from the predicted meshes to the ground-truth meshes, and vice versa for CD(gt, pred). 
  % The chamfer distances are multiplied by 10 for better display. 
  % Our results are consistently better than IGR over all metrics. 
  }
  \begin{tabular}{cccccc}
    \toprule
    Network      & CD$\downarrow$     & NC$\uparrow$    & CD(pred, gt)$\downarrow$   & CD(gt, pred) $\downarrow$  & Time$\downarrow$  \\
   \midrule
    IGR          & 0.499              & 0.875           & 0.854           & 0.143         & 110.8s  \\
    Ours         & \tbf{0.048}        & \tbf{0.964}     & \tbf{0.048}     & \tbf{0.047}   & \tbf{0.281s}  \\
    \bottomrule
  \end{tabular}
  % \vspace{-4mm}
  \label{tab:dfaust}
\end{table}

\paragraph{Results}
We conduct the comparison with IGR~\cite{Gropp2020} using the code provided by the authors.
In SALD~\cite{Atzmon2021}, IGR has already been thoroughly compared with SAL~\cite{Atzmon2020} and SALD on D-Faust: SAL cannot reconstruct geometry details compared with IGR, and SALD achieves a similar numerical performance to IGR. % (The code of SALD is not released).
The key difference between IGR and SAL/SALD is the loss function.
The recently proposed ACORN~\cite{Martel2021} requires the ground-truth volumetric field values as the training target and cannot be used in this task,  as the volumetric fields of ACORN are not continuous and the gradient is not well defined.
Thus, we mainly conduct the comparison with IGR, and the results are summarized in \cref{tab:dfaust}.
For evaluation, we report the Chamfer distance and normal consistency in \cref{subsec:shapenet}, as well as the one-side Chamfer distance following IGR.
Since the output meshes of IGR are often incomplete as shown in \cref{fig:dfaust}, the mIoU for IGR is not well defined, thus we omit this metric in this experiment.
According to \cref{tab:dfaust}, our results are clearly much better than IGR.
Especially, the Chamfer distance of our results is smaller than that of IGR by one order.
The visual comparisons in \cref{fig:dfaust} demonstrate that IGR has a poor ability to reconstruct geometric details from input point clouds, for example, the hands of the bodies are missing, as highlighted in the red box.
Additionally, the reconstructed shapes of IGR contain fake surface sheets on the feet, as shown in \cref{fig:dfaust}.

We need to mention that IGR is based on the autodecoder proposed by~\cite{Park2019}.
For each testing shape, the latent code has to be optimized separately by 800 iterations according to the setting of IGR, therefore IGR consumes a lot of time in the testing stage.
Our network can produce the output directly in a single forward pass.
We also record the average running time of producing one result from the testing dataset in \cref{tab:dfaust}.
Our method runs extremely fast and is 394 times faster than IGR on average.

\paragraph{Generalization Ability}
We observe that our network trained on D-Faust demonstrates strong generalization ability.
We directly apply the trained network to several point clouds from the Stanford 3D Scanning Repository and 2 scene-level point clouds from SceneNN~\cite{Handa2015}.
The results of surface reconstruction are shown in \cref{fig:surface} and \cref{fig:teaser}.
Previous state-of-the-art learning-based surface reconstruction methods, like SIREN~\cite{Sitzmann2020}, need several hours to train the network to get similar results, while our network generates these results by a single forward pass, which takes about \SI{420}{ms} on average for the meshes in \cref{fig:surface}.

\begin{figure}[t]
  \centering
  \begin{overpic}[width=\linewidth]{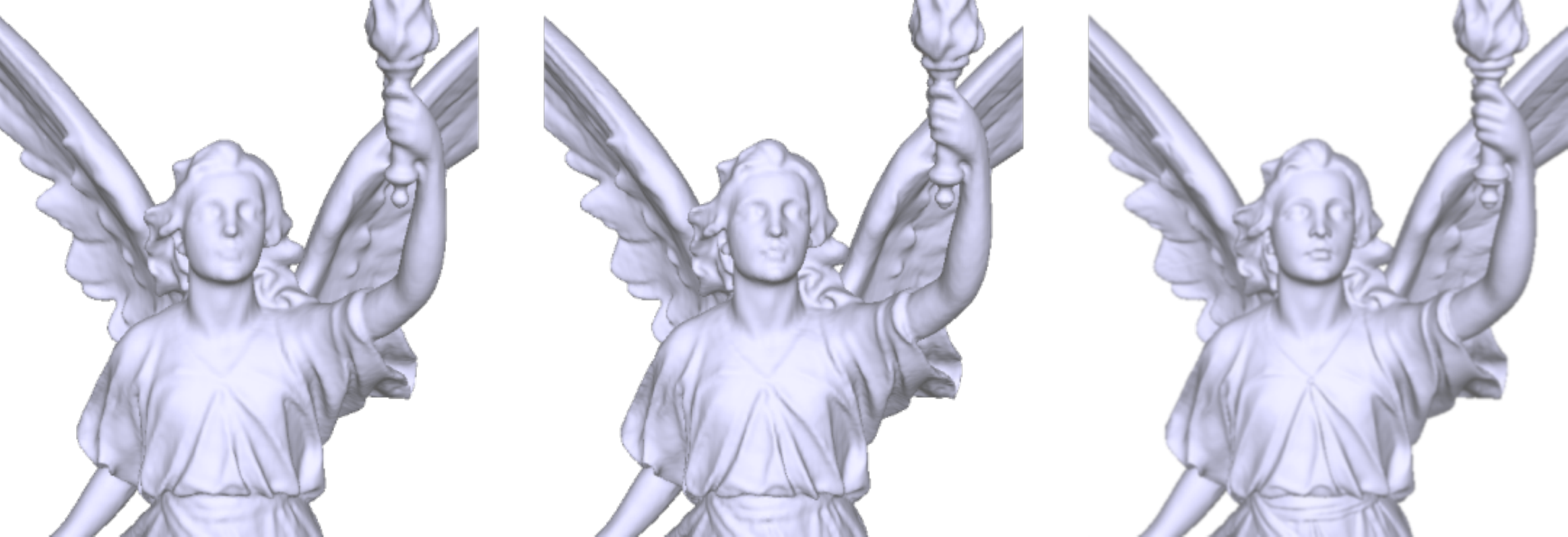}
    \put(4, -3){\small (a) Initial output}
    \put(38, -3){\small (b) After fintuning}
    \put(72, -3){\small (c) Ground-truth}
  \end{overpic}
  \caption{Improve the surface reconstruction with finetuning. (a) is the initial output of our pretrained network, (b) is the improved output after 2000 iterations of finetuning.
  The details near the mouth of Lucy are gradually recovered. 
  }
  \label{fig:finetune}
  \vspace{-4mm}
\end{figure}

\paragraph{Finetuning}
Given the pretrained weights, we can finetune the network to further improve the results.
In \cref{fig:finetune}, we finetuned the network with the same loss function by 2000 iterations with Adam and a learning rate of 0.0001.
More geometric features are reproduced gradually after finetuning.

\subsection{Autoencoder} \label{subsec:autoencoder}

In the previous sections, we conduct experiments with U-Net. In this section, we further verify the flexibility and effectiveness of our method with an autoencoder, which does not have skip connections between the encoder and the decoder.

\paragraph{Dataset}
We train an autoencoder with 13 categories of CAD shapes from ShapeNet~\cite{Chang2015} and used the same data splitting as IM-Net~\cite{Chen2019}: 35019 shapes for training and 8762 shapes for testing.
The data preprocessing is the same as \cref{subsec:shapenet}, and the depth of the octree is set as 6.

\begin{figure}[t]
  \centering
  \begin{overpic}[width=\linewidth]{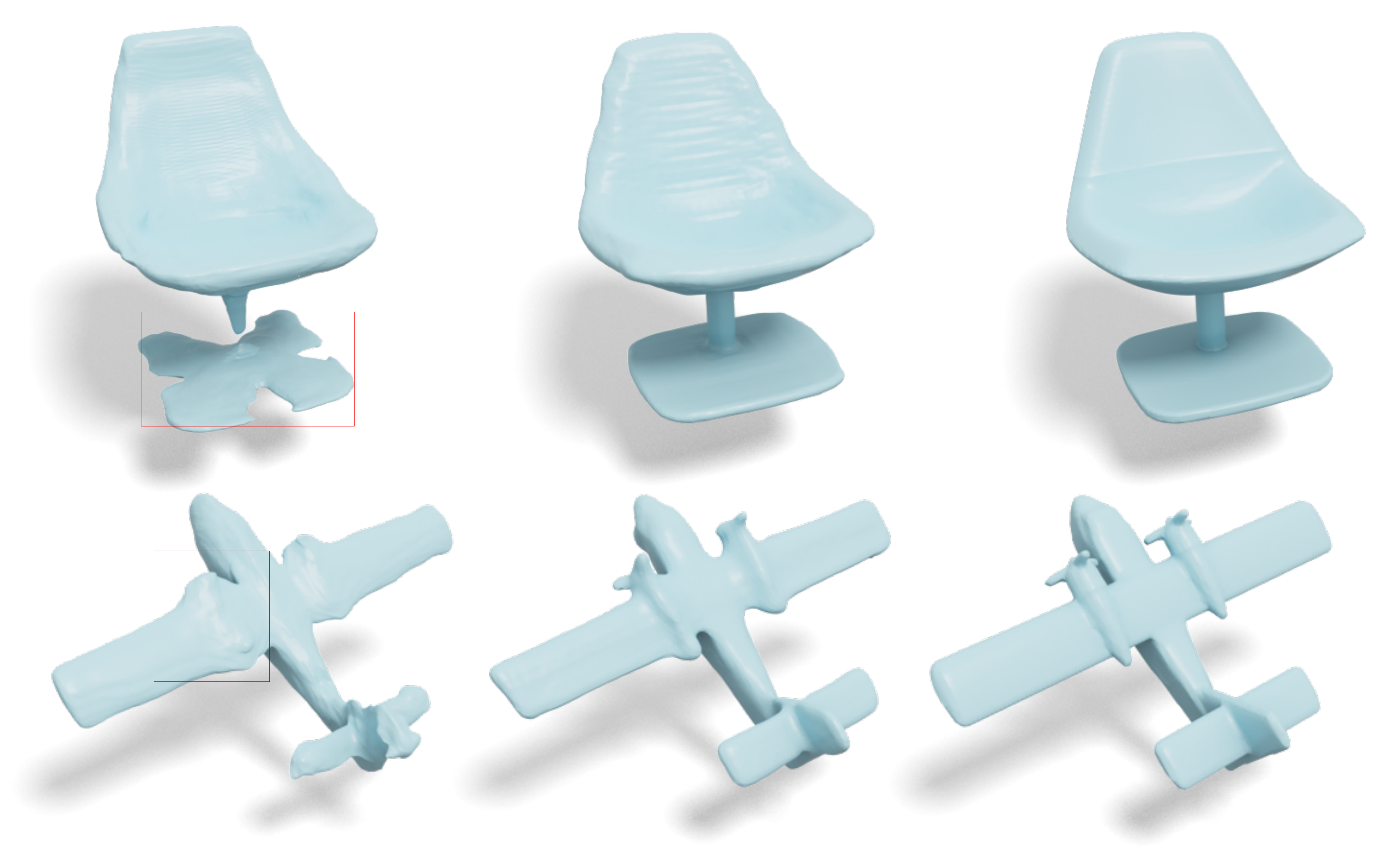}
    \put(13, -2){\small (a) IM-Net}
    \put(49, -2){\small (b) Ours}
    \put(74, -2){\small (c) Ground-truth}
  \end{overpic}
  \caption{Visual results of autoencoders. The results of IM-Net have artifacts and contain less  details as highlighted in the red boxes. }
  \label{fig:autoencoder}
   \vspace{-4mm}
\end{figure}

\begin{table}[t]
  \tablestyle{16pt}{1.1}
  \vspace{4pt}
  \caption{Quantitative evaluation of autoencoders on ShapeNet. 
    The Chamfer distance (CD) is multiplied by 100 for better display.}
  \begin{tabular}{cccc}
    \toprule
    Network      & CD$\downarrow$    & NC$\uparrow$       & Time$\downarrow$  \\
   \midrule
    AtlasNet     & 1.17              & 0.807              & -     \\
    OccNet       & 1.94              & 0.821              & -     \\
    IM-Net       & 1.37              & 0.811              & 0.841s \\
    Ours         & \tbf{0.89}        & \tbf{0.921}        & \tbf{0.143}s \\
    \bottomrule
  \end{tabular}
  \vspace{-2mm}
  \label{tab:autoencoder}
\end{table}

\paragraph{Results}
We use the same training settings and loss function as \cref{subsec:shapenet}.
We report the evaluation metrics in \cref{tab:autoencoder}.
The results of AtlasNet~\cite{Groueix2018} and OccNet~\cite{Mescheder2019} are provided by the authors of IM-Net~\cite{Chen2019}, and we use the code of IM-Net to generate its results.
In this experiment, the Chamfer distance and normal consistency are calculated with 4096 sample points following IM-Net.
The encoders of both IM-Net and OccNet are a volumetric CNN, and their decoders are a coordinate-based MLP.
The network of AtlasNet is a point-based autoencoder.
And all networks are trained on 13 categories.
It is clear that our performance is significantly better than IM-Net, OccNet, and AtlasNet.
The results further verify the effectiveness of our network over coordinate-based MLPs in learning volumetric fields.
For  running time, we record the average time of a network forward pass and query the field values of $100^3$ uniformly sampled points for each shape in the testing dataset.
As shown in the last column of \cref{tab:autoencoder}, our method is 5.9 times faster than IM-Net.
The visual comparison is shown in \cref{fig:autoencoder}.
Compared with IM-Net, our results are more similar to ground truth, while IM-Net cannot faithfully reconstruct the base of the chair and engine of the airplane.

% \paragraph{Latent GAN}
% After training the autoencoder, we extract the latent code and train a latent GAN~\cite{Achlioptas2018}.
% We show the random samples generated by the latent GAN in \todo{Figure} and shape interpolation in \todo{Figure}.
% It can be seen that the samples are of good quality, and the interpolation results demonstrate smooth transitions even between shapes of different categories.

\section{Conclusion} \label{sec:conclusion}

In this paper, we propose dual octree-based graph neural networks to learn adaptive volumetric shape representations.
We use an octree to partition the 3D volume into adaptive voxels and build a dual graph to connect adjacent voxels.
We propose a novel graph convolution, which operates on voxels with multiple resolutions, and construct encoder and decoder networks to extract adaptive features.
We blend these features with a novel neural MPU module to generate continuous volumetric fields.
Our method demonstrates great efficiency and achieves state-of-the-art performance in a series of shape reconstruction tasks.

In our work, the following aspects are not fully investigated, and we would like to explore them further.
\paragraph{Concurrent Downsampling and Upsampling on Leaf Nodes}
In our current implementation, at one time, only the octree nodes at one level are merged during graph downsampling, or subdivided during graph upsampling. Another possible strategy is to run these operators on all leaf nodes in different levels simultaneously, so that the messages between nodes can be passed and aggregated via the adaptive hierarchy of the dual octree more effectively. We would like to explore whether this alternative strategy can further maximize the performance of our network.

\paragraph{Optimal Octree Structure}
Currently, the adaptiveness of the feature volume in the encoder and decoder is determined by the pre-built octree based on the input, which may not be optimal. It is possible to combine the octree optimization procedure proposed in ACORN~\cite{Martel2021} with our network to increase the compactness of the octree and further improve efficiency.

\paragraph{Shape Analysis}
Our graph convolution is general and can also be used in shape analysis.
We have built a ResNet with our graph convolution and tested it on the ModelNet40 classification~\cite{Wu2015}. The classification accuracy is 
promising: 92.4\%, although it does not outperform other state-of-the-art networks, such as DGCNN~\cite{Wang2019c} (92.9\%) or PCT~\cite{Guo2021} (93.2\%). In the future, we would like to explore more possibilities for applying our method to shape analysis, especially for shapes in volumetric representations.

%%%%%%%% ACKs
\begin{acks}
We thank the anonymous reviewers for their valuable feedback.
\end{acks}

%%%%%%%%% REFERENCE
\bibliographystyle{ACM-Reference-Format}
\bibliography{src/ref/reference}

% %%%%%%%%% APPENDIX
% \clearpage
% \appendix
% \input{src/appendix/appendix}

\end{document}